\documentclass{article}

\usepackage[preprint]{neurips_2026}
\usepackage[utf8]{inputenc}
\usepackage[T1]{fontenc}
\usepackage{hyperref}
\usepackage{url}
\usepackage{booktabs}
\usepackage{amsfonts}
\usepackage{amsmath,amssymb}
\usepackage{microtype}
\usepackage{xcolor}
\usepackage{graphicx}
\usepackage{mathtools}
\usepackage{multirow}
\usepackage{siunitx}
\usepackage{array}
\usepackage{caption}
\usepackage{subcaption}
\usepackage{threeparttable}
\usepackage{enumitem}
\usepackage{placeins}
\usepackage{float}
\usepackage{tikz}
\usepackage{algorithm}
\usepackage{algpseudocode}
\usetikzlibrary{arrows.meta,positioning,fit,backgrounds,calc}

\definecolor{bestred}{RGB}{180,20,20}
\definecolor{secondblue}{RGB}{25,85,180}
\newcommand{\best}[1]{\textbf{\textcolor{bestred}{#1}}}
\newcommand{\second}[1]{\textbf{\textcolor{secondblue}{#1}}}

\title{FALCON-Discover: Discovering Concentrated False-Confidence Regions for Calibration}

\author{
Filippo Cenacchi, Longbing Cao, and Runze Yang\\
Macquarie University, Sydney, Australia\\
\texttt{filippo.cenacchi@mq.edu.au, longbing.cao@mq.edu.au, runze.yang@hdr.mq.edu.au}
}

\begin{document}
\maketitle
\vspace{-8pt}

\begin{abstract}
Calibration is usually evaluated at the aggregate level: confidence is judged by whether it aligns, on average across samples, with empirical correctness. This perspective is indispensable but incomplete, because the most operationally dangerous failures are often not average errors but a relatively small subset of predictions that remain highly confident despite being wrong. We study this failure mode as its own empirical object and introduce \emph{false-confidence concentration}: the extent to which dangerous confident errors occupy a compact and discoverable slice of the prediction space. This shifts the focus from population-level calibration quality alone to the discovery of sample-level regions in which confidence becomes systematically unreliable. We present \emph{FALCON-Discover}, a post-hoc discrepancy-discovery framework that ranks predictions using signals derived from confidence, local support, neighborhood agreement, and perturbation stability under controlled support-preserving perturbations. The framework is model-agnostic and does not alter the underlying classifier. Across seven heterogeneous binary tabular datasets, with four seeds, five-fold cross-fitting, and strong base learners including XGBoost and CatBoost, we find that false-confidence concentration is a recurrent but regime-dependent property. At the main confidence threshold, the discrepancy family substantially outperforms the strongest validation-selected prior baseline in the strongest regimes, while raw confidence ranking recovers little dangerous-error mass. The strongest family member varies: learned discrepancy is strongest when multiple cues must be combined, whereas stability-centered ranking is strongest when local decisional fragility dominates. These results establish false-confidence concentration as a family-level discovery problem rather than a single-score dominance claim, and motivate calibration strategies that explicitly target regions where confidence, support, and stability diverge.
\end{abstract}

\section{Introduction}

A deployed predictive model is not consumed through aggregate statistics alone. In practice, people encounter one prediction, one confidence value, and one decision consequence at a time. This simple fact creates a persistent tension in calibration research. On one hand, aggregate calibration summaries such as Expected Calibration Error, Brier score, negative log-likelihood, and reliability diagrams remain indispensable because they answer whether predicted confidence is aligned, at the population level, with observed correctness on average \citep{guo2017calibration,platt1999probabilistic,zadrozny2001obtaining,zadrozny2002transforming,vaeicenavicius2019evaluating,minderer2021revisiting}. On the other hand, these summaries do not directly answer the practically decisive question: \emph{which predictions are dangerous to trust?} A model can appear well calibrated in aggregate while still containing a small but consequential subset of predictions that remain highly confident despite being wrong. These are often the failures that matter most because they are least likely to be inspected, least likely to trigger caution, but most likely to propagate harmful downstream decisions. The post-hoc calibration literature has made major progress on score alignment, but it has largely retained a scalar view of the problem. Temperature scaling \citep{guo2017calibration}, Platt scaling \citep{platt1999probabilistic}, isotonic regression \citep{zadrozny2001obtaining,zadrozny2002transforming}, beta calibration \citep{kull2017beta,kull2019beyond}, Bayesian binning \citep{naeini2015obtaining}, and Dirichlet calibration all focus on improving a mapping from an existing score to a better calibrated probability estimate. This body of work shows that strong discrimination does not imply strong calibration and that probability quality can often be improved substantially after training \citep{niculescu2005predicting,kuleshov2018accurate}. Recent work has also expanded the calibration landscape beyond the classical post-hoc setting, including train-time uncertainty--error alignment \citep{mendes2025clue}, confidence--calibration dynamics during model training \citep{durai2025phases}, and post-hoc methods designed to reduce confidently incorrect predictions beyond standard aggregate calibration metrics \citep{denoodt2025variance,gharoun2025uncertainty}. Yet even in these newer directions, the central object usually remains a score distribution summarized globally, rather than a compact structural slice in which dangerous confident failures accumulate \citep{denoodt2025variance,mendes2025clue,durai2025phases,gharoun2025uncertainty}. A neighboring literature goes further by asking whether a model’s output should be trusted, whether a prediction should be abstained from, or whether an auxiliary score predicts correctness more effectively than native confidence \citep{jiang2018trustscore,corbiere2019addressing,hendrycks2017baseline,geifman2017selective,elyaniv2010foundations,geifman2019selectivenet,romano2020classification}. These directions are highly relevant because they restore decision-time risk to the reliability problem. However, even here the dominant framing remains mostly score-centric: \textbf{can one construct a better trust score, a better correctness predictor, or a better abstention signal than confidence alone?} This paper starts from a different observation: dangerous overconfidence is often concentrated, so a small subset of samples can contain a disproportionate share of all high-confidence errors. This suggests a more useful empirical target: a compact, structurally identifiable slice of the prediction space whose review surfaces many failures that are unsafe to trust. We therefore reframe calibration analysis as discrepancy discovery: rather than asking only whether confidence can be globally reshaped, we ask whether predictions ranked by disagreement between confidence, support, and structural stability recover more dangerous error mass than confidence alone. This yields \emph{FALCON-Discover}, a post-hoc, model-agnostic framework that builds a discrepancy representation from native decision signals, local support, neighborhood agreement, and perturbation stability, then uses it to rank dangerous errors, localize discrepancy regions, and derive calibration-facing weights. Empirically, false-confidence concentration emerges as a regime-resolved phenomenon: it is strong and threshold-stable in several benchmark regimes, mixed when stronger predictors absorb local structure, and boundary-limited when high-confidence error events become sparse. The strongest detector varies by regime: the learned discrepancy ranker is strongest in some cases, while a simpler stability-centered rule is stronger in others, showing that the phenomenon is broader than any single detector. Accordingly, ranking is only one operational view; the broader contribution is a discrepancy representation for diagnosis, localization, and calibration refinement. Our goal is not to replace aggregate calibration analysis, but to complement it with a sample-level view of where confident predictions become structurally unsafe to trust.  We evaluate this claim on seven heterogeneous binary tabular datasets using four seeds and five-fold cross-fitting, fixing $\tau=0.90$ and review budgets in $\{5\%,10\%,15\%,20\%\}$, and measuring concentration via FalseConf-AUROC and Capture@20 against validation-selected calibration and trust-scoring baselines.

The paper makes five contributions. First, it formalizes \emph{false-confidence concentration} as a structural reliability object: the recoverability of compact high-risk error slices hidden by aggregate calibration summaries. Second, it introduces \emph{prediction-to-structure learning}, a post-hoc paradigm in which held-out prediction behavior is lifted into a reusable discrepancy state rather than reduced to another scalar trust score. Third, it characterizes when concentration should arise: high native certainty must coincide with weak local evidence or instability strongly enough to amplify dangerous-error density inside a compact conflict set. Fourth, it supplies controlled multi-dataset evidence across thresholds and strong tabular backbones. Fifth, it separates strong, mixed, and boundary regimes, making the claim explicitly family-level rather than a universal dominance statement.

\section{Related Work}

Post-hoc calibration is usually framed as aligning a scalar score with empirical correctness. Temperature scaling remains a strong standard baseline because it is simple and architecture-agnostic \citep{guo2017calibration}. Platt scaling, isotonic regression, beta calibration, Bayesian binning, and Dirichlet calibration occupy nearby points in the same design space, transforming raw scores into probability estimates that better match observed frequencies \citep{platt1999probabilistic,zadrozny2001obtaining,zadrozny2002transforming,naeini2015obtaining,kull2017beta,kull2019beyond}. The broader lesson is that discrimination and calibration are distinct, and that accurate classifiers can still be poor sources of probabilities \citep{niculescu2005predicting,kuleshov2018accurate,minderer2021revisiting}. Recent work has extended this view through efficient post-hoc uncertainty calibration, train-time uncertainty--error alignment, and confidence--calibration dynamics in modern language models \citep{denoodt2025variance,mendes2025clue,durai2025phases}. Our work addresses a complementary question left underemphasized by this literature: where dangerous confident failures accumulate, how concentrated they are, and whether a small review slice can recover them efficiently. A nearby literature asks whether a model’s output should be trusted. TrustScore replaces raw confidence with a class-relative support signal \citep{jiang2018trustscore}; failure-prediction methods learn whether a model is likely to be wrong \citep{corbiere2019addressing,hendrycks2017baseline}; and selective classification methods study reject, defer, or bounded-error prediction under partial coverage \citep{elyaniv2010foundations,geifman2017selective,geifman2019selectivenet,romano2020classification}. A broader uncertainty literature studies reliability under distribution shift, ensemble disagreement, Bayesian approximations, and energy-based formulations \citep{lakshminarayanan2017simple,gal2016dropout,ovadia2019can,ashukha2020pitfalls,liu2020energy}, while conformal prediction addresses set-valued uncertainty with coverage guarantees \citep{vovk2005algorithmic,romano2020classification}. Structured reporting frameworks such as Model Cards and Datasheets add the complementary principle that failure modes should be made legible rather than hidden behind averages \citep{mitchell2019model,gebru2021datasheets}. These lines of work reinforce the same methodological intuition: confidence should not be interpreted in isolation. What remains underdeveloped is a direct study of whether \emph{confident} errors form recoverable local failure regions, whether those regions can be surfaced by reusable discrepancy states rather than by another calibrated probability or reject score, and whether the resulting structure can guide later calibration refinement. This is the gap addressed.

\section{Method}

\begin{figure}[H]
\centering
\includegraphics[width=0.90\textwidth]{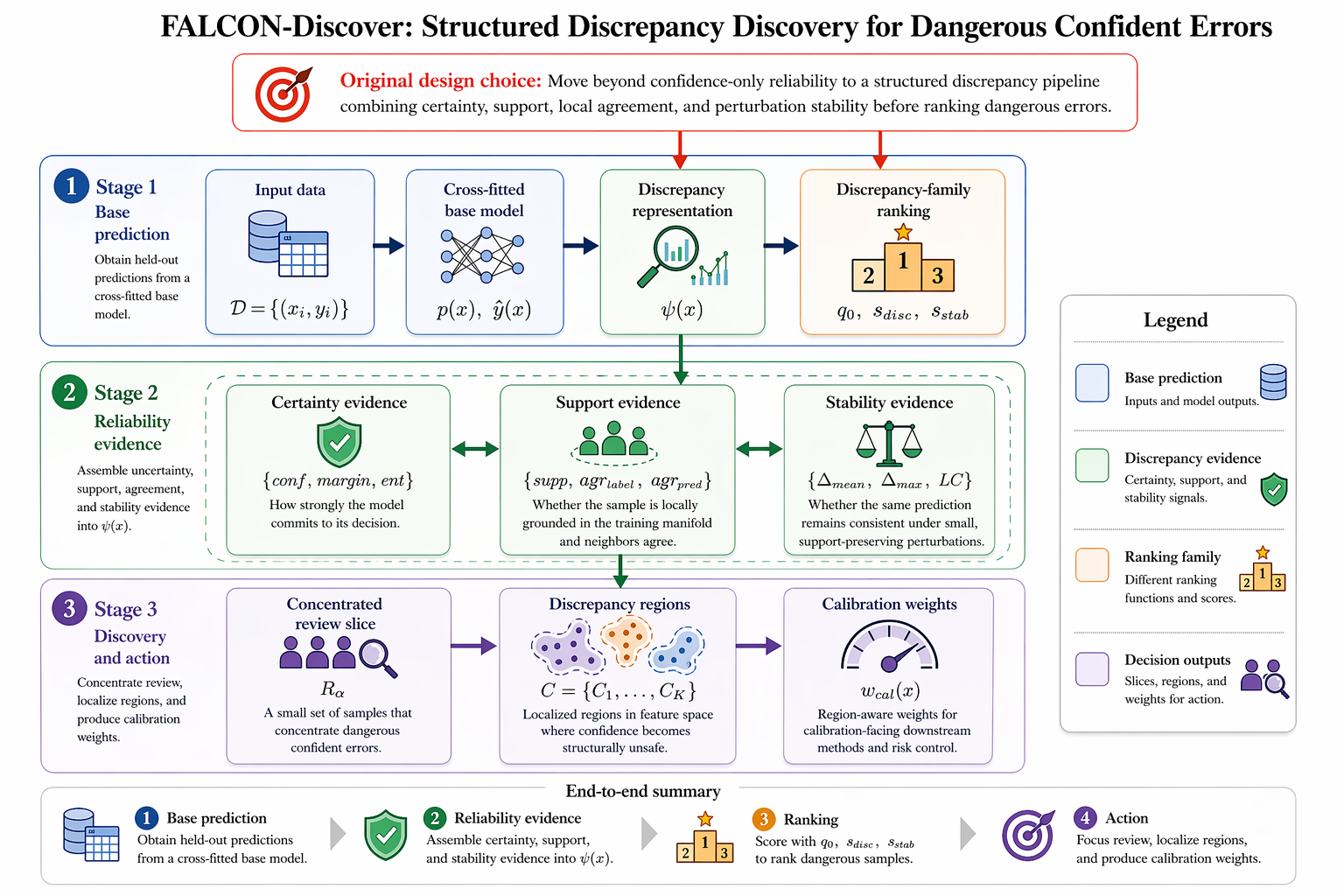}
\caption{\textbf{Overview of FALCON-Discover.} Held-out predictions from a cross-fitted base model are mapped to a discrepancy representation $\psi(x)$ combining certainty, local support/agreement, and perturbation stability, then reused to rank dangerous samples, localize discrepancy regions, and derive calibration-facing weights.}
\label{fig:falcon_design}
\end{figure}

Figure~\ref{fig:falcon_design} summarizes the FALCON-Discover pipeline. The method is motivated by a simple hypothesis: dangerous overconfidence should be most visible when strong predictive certainty coexists with weak local support or instability under small, support-preserving changes. FALCON-Discover therefore does not introduce one more scalar trust score in isolation. Instead, it decomposes reliability into three complementary questions for each sample: how confident the model is, how well that sample is supported by nearby training data, and how stable the same prediction remains under small local perturbations. These signals are combined into a shared discrepancy representation, which is then reused to rank dangerous samples, localize discrepancy regions, and derive calibration-facing weights. The central design choice is thus a structured discrepancy representation whose role is diagnostic and actionable rather than a collection of disconnected engineered scores. The learning object is not a new classifier, but a post-hoc \emph{failure-discovery representation}: the base model is fixed, while held-out behavior is mapped into a discrepancy space that exposes where confidence, support, and stability conflict. This distinguishes FALCON-Discover from calibration maps, failure predictors, and abstention rules, which usually return a probability, correctness score, or reject decision rather than a reusable structure for ranking, localization, and calibration guidance.

FALCON-Discover is instantiated for binary classification, matching all benchmark tasks. Stage~1 produces held-out scores $p(x)$ and decisions $\hat{y}(x)$; Stage~2 builds $\psi(x)$ from certainty, support/agreement, and stability signals; and Stage~3 reuses $\psi(x)$ for ranking, region discovery, and calibration-facing weights. Multiclass extension is plausible but outside the present scope. Let a trained binary classifier produce a positive-class score $p(x)\in[0,1]$ for sample $x$, with hard decision
\begin{equation}
\tiny
\hat{y}(x)=\mathbb{I}[p(x)\ge 0.5].
\label{eq:hard_prediction}
\end{equation}
We define confidence as
\begin{equation}
\tiny
\mathrm{conf}(x)=\max\{p(x),1-p(x)\}.
\label{eq:confidence}
\end{equation}
For a user-selected confidence threshold $\tau\in(0,1)$, we define the false-confidence event
\begin{equation}
\tiny
\mathrm{FC}_{\tau}(x)
=
\mathbb{I}\!\left[
\underbrace{\hat{y}(x)\neq y}_{\text{prediction is wrong}}
\;\land\;
\underbrace{\mathrm{conf}(x)\ge\tau}_{\text{model is highly confident}}
\right].
\label{eq:false_confidence_event}
\end{equation}
$\mathrm{FC}_{\tau}(x)$ isolates the failure cases most likely to evade manual skepticism because they combine incorrectness with high apparent certainty. Given a ranking rule $s(x)$ and a review budget $\alpha\in(0,1)$, let $\mathrm{Top}_{\alpha}(s)$ denote the top $\alpha$ fraction of test samples sorted by descending $s(x)$. We evaluate concentration through
{\scriptsize
\begin{equation}
\mathrm{Capture}@\alpha(s)
=
\frac{\sum_{x\in \mathrm{Top}_{\alpha}(s)} \mathrm{FC}_{\tau}(x)}
     {\sum_{x} \mathrm{FC}_{\tau}(x)} .
\label{eq:capture}
\end{equation}
}
Equation~\eqref{eq:capture} formalizes the paper’s operational target: under a fixed review budget, how much dangerous error mass can be surfaced? To state more precisely when false-confidence concentration should arise, define the \emph{conflict-amplification gap}
{\scriptsize
\begin{equation}
\Gamma_{\tau}(c,e,s)
=
\Pr\!\big(\mathrm{FC}_{\tau}(x)=1 \mid \mathrm{conf}(x)\ge c,\ \mathcal{E}(x)\le e,\ \mathcal{S}(x)\ge s\big)
-
\Pr\!\big(\mathrm{FC}_{\tau}(x)=1 \mid \mathrm{conf}(x)\ge c\big),
\label{eq:conflict_gap}
\end{equation}
}
together with the corresponding slice mass
{\scriptsize
\begin{equation}
M(c,e,s)
=
\Pr\!\big(\mathrm{conf}(x)\ge c,\ \mathcal{E}(x)\le e,\ \mathcal{S}(x)\ge s\big).
\label{eq:slice_mass}
\end{equation}
}
Here $\mathcal{E}(x)$ denotes local evidence quantities such as support or neighborhood agreement, and $\mathcal{S}(x)$ denotes instability quantities such as $\Delta_{\mathrm{mean}}(x)$ or $1-\mathrm{LC}(x)$. False-confidence concentration should be strongest when $\Gamma_{\tau}(c,e,s)$ is clearly positive on a slice whose mass is non-trivial but not diffuse: in that case, conditioning on structural conflict raises dangerous-error density relative to high confidence alone while keeping the region compact enough to be recoverable under a small review budget. By contrast, when confidence, support, and stability remain aligned, $\Gamma_{\tau}(c,e,s)$ becomes small and discrepancy-aware ranking should collapse toward confidence ranking. This yields the three empirical regimes used later in the paper: \emph{strong} regimes exhibit clear conflict amplification on recoverable slices, \emph{mixed} regimes exhibit weaker or backbone-dependent amplification, and \emph{boundary} regimes have event spaces too sparse for reliable concentration estimates. The characterization places the paper between global calibration and selective prediction: calibration asks whether confidence matches frequency on average \citep{guo2017calibration,vaeicenavicius2019evaluating,minderer2021revisiting}, while trust scoring and selective prediction ask for a better accept/reject signal \citep{jiang2018trustscore,elyaniv2010foundations,romano2020classification}; our target is the intermediate structural question of whether dangerous high-confidence error mass occupies a compact conflict region at all. The contribution is therefore structural rather than distribution-free: without additional assumptions linking event prevalence, conflict amplification, and slice mass, no non-vacuous guarantee can identify concentrated false-confidence regions from finite held-out data alone. We therefore treat concentration as a measurable property of a trained model and dataset, and evaluate it through held-out ranking, threshold robustness, bootstrap intervals, fixed-backbone sensitivity, and regime analysis. The method then instantiates a family of discrepancy-aware ranking rules and tests whether a small ranked slice captures a disproportionate share of false-confidence events. In this sense, the design is narrower than an end-to-end abstention architecture and more operational than a purely global calibration map, because it targets concentrated reliability failure directly and translates the discovered structure into region summaries and calibration-facing weights. Strict cross-fitting is essential. Let $\mathcal{D}_{\mathrm{train}}$, $\mathcal{D}_{\mathrm{val}}$, and $\mathcal{D}_{\mathrm{test}}$ denote disjoint train, validation, and test partitions. For the current benchmark, the base classifier is selected from a candidate library containing logistic regression, random forests, extra trees, HistGradientBoosting, multilayer perceptrons, XGBoost, and CatBoost. For each candidate in this pool, we produce out-of-fold predictions on $\mathcal{D}_{\mathrm{train}}$ using stratified $K$-fold cross-fitting and select the model with the strongest out-of-fold label AUROC rather than raw accuracy. The chosen base learner is then refit on the full training partition and evaluated on validation and test splits. All downstream discrepancy features are therefore constructed from out-of-fold or held-out predictions, not from predictions contaminated by training leakage. The three signal families play distinct roles: certainty captures how strongly the model commits to its decision, support captures whether that decision is locally grounded in the training data, and stability captures whether the same decision persists under small local changes that preserve coarse semantic identity. For each sample $x$, we construct a structured signal family $\psi(x)$ from these three signal families. The first family captures native certainty:
\begin{equation}
\tiny
\mathrm{margin}(x)=|2p(x)-1|,
\label{eq:margin}
\end{equation}
\begin{equation}
\tiny
\mathrm{ent}(x)=
-
p(x)\log p(x)
-
(1-p(x))\log(1-p(x)).
\label{eq:entropy}
\end{equation}
The second family captures support and local agreement. Let $\tilde{x}$ denote the transformed and standardized representation of $x$. We define a normalized support score
\begin{equation}
\tiny
\mathrm{supp}(x)
=
1-\mathrm{clip}\!\left(
\frac{d_{\mathrm{sup}}(x)-q_{0.50}}{q_{0.95}-q_{0.50}},
\,0,\,1
\right),
\label{eq:support}
\end{equation}
where
\begin{equation}
\tiny
d_{\mathrm{sup}}(x)
=
\underbrace{\frac{1}{d}\sum_{j=1}^{d}\frac{(\tilde{x}_j-\mu_j)^2}{\sigma_j^2+\varepsilon}}_{\text{normalized distance from the transformed training manifold}}
\label{eq:support_distance}
\end{equation}
and $q_{0.50},q_{0.95}$ are the training-set median and upper quantile. We also define neighborhood agreement in transformed space. Let $\mathcal{N}_k(x)$ be the $k$ nearest neighbors of $x$ among the cross-fitted training embeddings. Then
\begin{equation}
\tiny
\mathrm{agr}_{\mathrm{label}}(x)
=
\max\!\left\{
\frac{1}{k}\sum_{x'\in\mathcal{N}_k(x)}\mathbb{I}[y(x')=1],\,
\frac{1}{k}\sum_{x'\in\mathcal{N}_k(x)}\mathbb{I}[y(x')=0]
\right\},
\label{eq:label_agreement}
\end{equation}

The third family captures perturbation stability. A central innovation of the framework is to treat local instability as a first-class component of dangerous overconfidence. For each transformed sample $\tilde{x}$, we generate a perturbation set $\mathcal{P}(x)$ by partially mixing it with nearby support neighbors:
\begin{equation}
\tiny
\tilde{x}^{\,\prime}
=
(1-\lambda)\tilde{x}
+
\lambda \tilde{x}_{\mathrm{nbr}},
\qquad
\lambda\in\{0.10,0.20,0.30\}.
\label{eq:perturbation}
\end{equation}
These perturbations preserve local identity while probing prediction stability. We then define
\begin{equation}
\tiny
\Delta_{\mathrm{mean}}(x)
=
\frac{1}{|\mathcal{P}(x)|}
\sum_{x'\in\mathcal{P}(x)}
|p(x')-p(x)|,
\label{eq:mean_drift}
\end{equation}
\begin{equation}
\tiny
\Delta_{\mathrm{max}}(x)
=
\max_{x'\in\mathcal{P}(x)}
|p(x')-p(x)|,
\label{eq:max_drift}
\end{equation}
\begin{equation}
\tiny
\mathrm{LC}(x)
=
\frac{1}{|\mathcal{P}(x)|}
\sum_{x'\in\mathcal{P}(x)}
\mathbb{I}[\hat{y}(x')=\hat{y}(x)],
\label{eq:label_consistency}
\end{equation}
\begin{equation}
\tiny
\mathrm{Var}_{\mathrm{logit}}(x)
=
\mathrm{Var}\!\left(
\log\frac{p(x')}{1-p(x')}
\right)_{x'\in\mathcal{P}(x)}.
\label{eq:logit_var}
\end{equation}
The final structured signal family is
\begin{equation}
\tiny
\begin{aligned}
\psi(x)
={}&\big[
\underbrace{\mathrm{conf}(x),\mathrm{margin}(x),\mathrm{ent}(x)}_{\text{native certainty}},\\
&\underbrace{\mathrm{supp}(x),\mathrm{agr}_{\mathrm{label}}(x),\mathrm{agr}_{\mathrm{pred}}(x)}_{\text{local support and agreement}},\\
&\underbrace{\Delta_{\mathrm{mean}}(x),\Delta_{\mathrm{max}}(x),\mathrm{LC}(x),\mathrm{Var}_{\mathrm{logit}}(x)}_{\text{structural stability}}
\big].
\end{aligned}
\label{eq:psi}
\end{equation}

Equation~\eqref{eq:psi} is the key organizational step of the method. The paper does not treat reliability as a scalar attribute attached to a prediction, but as a \emph{structured conflict state} in which certainty, support, agreement, and stability can align or disagree. This is the paper's learning paradigm: \emph{prediction-to-structure learning}, where held-out prediction behavior is first mapped into a discrepancy state and only then queried through ranking, localization, or calibration-facing actions. The representation is therefore the claim-bearing object; the downstream detectors are operational views of the same state rather than independent engineered scores. This framing also clarifies why concentration is regime-dependent. In certainty-dominated regimes the upper tail is initiated by high confidence and then sharpened by structural evidence; in stability-dominated regimes local perturbation sensitivity becomes the main separating cue; and in boundary regimes sparse false-confidence events weaken all views simultaneously. The scientific question is therefore not whether one detector is universally best, but whether concentrated failure structure is recoverable once prediction behavior is lifted into a discrepancy state. We evaluate random ranking and confidence-only ranking as minimal baselines. The stability-centered rule is intentionally simple and analytic: the larger coefficient is placed on average probability drift because sustained score movement is a stronger indicator of local unreliability than a single label flip alone, while label inconsistency remains a secondary corroborating signal. We then define a stability-centered score:
\begin{equation}
\tiny
s_{\mathrm{stab}}(x)
=
\underbrace{0.7\,\Delta_{\mathrm{mean}}(x)}_{\text{probability drift}}
+
\underbrace{0.3\,(1-\mathrm{LC}(x))}_{\text{label inconsistency}}.
\label{eq:stability_score}
\end{equation}
and a fixed analytic discrepancy score built from a coarse priority ladder,
{\tiny
\begin{align}
s_{\mathrm{disc}}(x)
&=
\underbrace{1.30\,\mathrm{conf}(x)}_{\text{high native certainty}}
+
\underbrace{1.00\,\Delta_{\mathrm{mean}}(x)}_{\text{average instability}}
+
\underbrace{0.80\,\Delta_{\mathrm{max}}(x)}_{\text{worst-case instability}}
\nonumber\\
&\quad+
\underbrace{1.00\,(1-\mathrm{LC}(x))}_{\text{label inconsistency}}
+
\underbrace{0.80\,(1-\mathrm{supp}(x))}_{\text{support collapse}}
+
\underbrace{0.70\,(1-\mathrm{agr}_{\mathrm{label}}(x))}_{\text{weak local agreement}}.
\label{eq:analytic_discrepancy}
\end{align}
}

Equation~\eqref{eq:analytic_discrepancy} is not intended as the paper’s uniquely best detector. Its role is methodological and falsification-oriented. If false-confidence concentration were visible only under a trained detector, the claim would be weaker because it could be attributed to detector flexibility rather than to the structure of the discrepancy state itself. By contrast, if the same phenomenon is already exposed by a coarse monotone aggregation of conflict signals, the claim is stronger: the state is informative before optimization. For that reason the coefficients are ordinal rather than tuned. They encode a transparent priority ladder among primary and corroborating conflict cues while deliberately avoiding the stronger claim that one fixed numeric weighting is universally optimal. Positive affine rescalings are immaterial, and the paper does not interpret the absolute value of $s_{\mathrm{disc}}(x)$; only the induced ranking matters. The empirical burden therefore lies on the state-level phenomenon and the family-level pattern, not on Equation~\eqref{eq:analytic_discrepancy} as a uniquely privileged formula. The learned discrepancy ranker is likewise introduced as a \emph{witness detector} over the standardized discrepancy state rather than as the paper’s primary source of capacity. The methodological claim is that held-out prediction behavior can be lifted into a structured conflict state whose separating information is already present before any powerful detector is applied. For that reason we intentionally use a weighted linear logit as the smallest transparent learned rule: if concentration is real, a minimal detector should expose it; if it is absent, expressive nonlinear fitting should not be allowed to manufacture it. The scientific burden is therefore placed on $\tilde{\psi}(x)$ rather than on detector complexity. This clarifies the logic of the discrepancy family: the fixed rules act as analytic witness scores, while the learned detector acts as a data-adaptive witness score over the same state. If these views agree in the strongest regimes, the evidence favors the discrepancy state itself rather than any one detector implementation. The paper’s methodological novelty therefore lies in the structured state and the prediction-to-structure learning paradigm, not in proposing a new high-capacity classifier.
\begin{equation}
\scriptsize
q_{\theta}(x)
=
\sigma\!\left(
\underbrace{w^{\top}\tilde{\psi}(x)+b}_{\text{learned combination of discrepancy signals}}
\right).
\label{eq:learned_ranker}
\end{equation}

with weighted objective

{\scriptsize
\begin{align}
\mathcal{L}(\theta)
&=
-
\sum_{i=1}^{n}
\underbrace{\omega_i}_{\text{false-confidence emphasis}}
\Big[
z_i\log q_{\theta}(x_i)
+
(1-z_i)\log(1-q_{\theta}(x_i))
\Big],
\label{eq:weighted_loss}
\end{align}
}

where $z_i=\mathrm{FC}_{\tau}(x_i)$ and

{\scriptsize
\begin{equation}
\omega_i
=
1+\eta\,\mathrm{FC}_{\tau}(x_i).
\label{eq:weights}
\end{equation}
}
This objective is fitted only on validation/calibration folds with observed labels; test labels are used exclusively for final reporting, so neither the learned ranker nor $w_{\mathrm{cal}}$ accesses test outcomes. To move from ranking to structure, we cluster the held-out discrepancy representation jointly with the learned discrepancy score, support, local agreement, and stability quantities; this localization step uses label-free discrepancy features and is never used to select the reported ranking rule. Finally, we derive a discrepancy-aware weighting profile for future calibration training. The coefficients are chosen to reflect an explicit priority ordering rather than numerical optimization: the profile should upweight samples most strongly when they are already false-confidence events, then when they are highly confident or locally unstable, and then when they show weaker but still informative structural warning signs such as weak support, weak neighborhood agreement, or label inconsistency. Because this profile is intended for later calibration refinement rather than for test-time prediction, its purpose is to allocate more training emphasis to the most operationally dangerous and structurally suspicious samples.
\begin{equation}
\tiny
w_{\mathrm{cal}}(x)
=
\underbrace{1.2\,\mathrm{conf}(x)}_{\text{high certainty}}
+
\underbrace{1.0\,(1-\mathrm{supp}(x))}_{\text{weak support}}
+
\underbrace{0.9\,(1-\mathrm{agr}_{\mathrm{label}}(x))}_{\text{weak local agreement}}
+
\underbrace{1.2\,\Delta_{\mathrm{mean}}(x)}_{\text{instability}}
+
\underbrace{1.0\,(1-\mathrm{LC}(x))}_{\text{label inconsistency}}
+
\underbrace{2.0\,\mathrm{FC}_{\tau}(x)}_{\text{false-confidence emphasis}}
\label{eq:cal_weight}
\end{equation}
The coefficient ordering in Equation~\eqref{eq:cal_weight} follows the operational logic of the method. The largest coefficient is assigned to $\mathrm{FC}_{\tau}(x)$ because samples that are already high-confidence errors are, by definition, the most urgent targets for later calibration refinement. The next-largest weights are assigned to $\mathrm{conf}(x)$ and $\Delta_{\mathrm{mean}}(x)$ because the target failure mode is specifically \emph{confident} error under local instability: high certainty makes the prediction dangerous to trust, while instability indicates that this certainty is poorly supported. The remaining terms receive slightly smaller but still positive weights because they act as corroborating structural cues. Weak support indicates that the sample lies in a poorly grounded region of the training manifold, weak neighborhood agreement indicates that nearby evidence is inconsistent, and label inconsistency captures local decisional fragility. The exact numeric values are therefore intended as transparent analytic priorities that encode this ordering, not as claims of universal optimality. The stronger empirical claim of the paper lies in the discrepancy representation and the family-level pattern it exposes, rather than in one uniquely best fixed weighting scheme. The novelty is therefore representational rather than architectural: familiar statistical signals are organized into a single discrepancy space that supports three operations usually treated separately, failure discovery, region localization, and calibration guidance.

\section{Experimental Design}

The evaluation targets concentration rather than global accuracy. We establish the phenomenon in seven public, heterogeneous binary tabular datasets (Adult, Bank Marketing, Spambase, MiniBooNE, Magic Telescope, Nomao, Phoneme), selected a priori because they provide a controlled but non-uniform testbed for the paper's structural claim: they vary in size, class balance, feature geometry, and false-confidence prevalence while preserving a clean binary setting in which confidence, local support, neighborhood agreement, and support-preserving perturbation stability can be defined transparently. This is not a convenience choice but a deliberate first laboratory for testing whether compact false-confidence slices exist before exporting the paradigm to multiclass, sequential, vision, language, or multimodal settings, where local evidence and perturbation semantics are harder to define cleanly. Strong tree-based models are included because recent tabular benchmarks identify them as highly competitive reference points \citep{grinsztajn2022tree,holzmuller2024better,mcelfresh2023when}. The final reported results aggregate across four seeds and use five-fold cross-fitting. The main threshold is fixed at $\tau=0.90$, and review budgets are fixed to $\{5\%,10\%,15\%,20\%\}$ for comparability. The prior baseline family consists of confidence-only ranking, temperature scaling, Platt scaling, isotonic calibration, beta calibration, and TrustScore, chosen a priori to span the canonical post-hoc calibration and trust-scoring families most directly relevant to the claim: score reshaping, non-parametric calibration, distributional calibration, confidence-only ranking, and local-support trust estimation. In addition, Appendix Table~\ref{tab:appendix_full_baselines} reports LightGBM, FT-Transformer, RealMLP, and SAINT comparator runs so that the discovery claim is tested not only against standard calibration baselines but also against strong modern tabular predictors. Experiments were run on a workstation with two NVIDIA RTX A6000 GPUs. For each dataset, seed, and threshold, prior baselines are tuned on the validation split only; the strongest validation baseline is then frozen and evaluated once on the held-out test split, preventing test-set cherry-picking and preserving the same information budget for all methods. Appendix Sections~\ref{sec:eval_reporting_details} and~\ref{sec:artifact_availability} summarize the reporting axes, metric families, success criterion, and anonymous review artifact used to evaluate and reproduce concentration beyond raw accuracy. Two evaluation choices deserve emphasis. First, the $20\%$ review slice reported in the main tables is an operationally interpretable reporting point rather than a cherry-picked budget: all operating-point summaries are computed over $\{5\%,10\%,15\%,20\%\}$, and the appendix visualizes the full budget sweep. Second, discrepancy regions are evaluated functionally rather than by raw cluster identity across seeds. Because cluster labels are non-identifiable up to permutation, stability is read through seed-aggregated region statistics---top-1 and top-2 false-confidence mass, mean support, mean instability, and correctness rate---rather than through literal cluster-label matching. For a discovery paper, the stable object is therefore the recurrence of the same \emph{type} of discrepancy-heavy slice, not an exact unsupervised label identity. Appendix Section~\ref{sec:functional_region_stability} makes this notion of region stability explicit.

\section{Results}

\begin{table}[t]
\centering

\tiny
\setlength{\tabcolsep}{4pt}
\renewcommand{\arraystretch}{1.08}
\caption{\textbf{Main multi-dataset result at $\tau=0.90$.} For each dataset we report the strongest \emph{validation-selected} prior baseline, the strongest discrepancy-family rule, and the resulting gains on the test split.}
\label{tab:main_discrepancy_family}
\begin{tabular}{l l l cc cc cc}
\toprule
\textbf{Dataset} & \textbf{Best prior} & \textbf{Best family} &
\textbf{Prior AUROC} & \textbf{Family AUROC} &
\textbf{Prior Cap@20} & \textbf{Family Cap@20} &
\textbf{$\Delta$AUROC} & \textbf{$\Delta$Cap@20} \\
\midrule
Adult              & trustscore      & learned discrepancy & 0.525 & \best{0.847} & 0.108 & \best{0.728} & \best{+0.322} & \best{+0.621} \\
Bank Marketing     & trustscore      & learned discrepancy & 0.629 & \best{0.859} & 0.212 & \best{0.740} & \best{+0.230} & \best{+0.528} \\
MiniBooNE          & trustscore      & learned discrepancy & 0.649 & \best{0.832} & 0.314 & \best{0.669} & \best{+0.182} & \best{+0.355} \\
Magic Telescope    & beta scaled     & learned discrepancy & 0.566 & \best{0.643} & 0.075 & \best{0.221} & \best{+0.077} & \best{+0.145} \\
Nomao              & trustscore      & stability           & 0.798 & \best{0.835} & 0.654 & \best{0.728} & \best{+0.038} & \best{+0.074} \\
Spambase           & trustscore      & stability           & 0.691 & \best{0.762} & \best{0.468} & 0.437 & \best{+0.071} & -0.031 \\
Phoneme            & isotonic scaled & learned discrepancy & \best{0.736} & 0.700 & \best{0.511} & 0.333 & -0.036 & -0.178 \\
\bottomrule
\end{tabular}
\end{table}

\begin{table}[t]
\centering
\tiny
\setlength{\tabcolsep}{2.6pt}
\renewcommand{\arraystretch}{0.96}

\begin{minipage}[t]{0.50\textwidth}
\centering
\caption{\textbf{Operational impact and threshold robustness.} Counts are means across seeds at $\tau=0.90$; last columns report $\Delta$\texttt{Cap@20}.}
\label{tab:operational_robustness}
\begin{tabular}{lcccccc}
\toprule
\textbf{Dataset} & \textbf{FC} & \textbf{Prior@20} & \textbf{Fam.@20} & \textbf{@.85} & \textbf{@.90} & \textbf{@.95} \\
\midrule
Adult & 117.0 & 11.25 & 84.00 & \best{+.610} & \best{+.621} & \best{+.484} \\
Bank & 99.0 & 21.00 & 73.25 & \best{+.497} & \best{+.528} & \best{+.464} \\
MiniBooNE & 268.5 & 84.25 & 179.50 & \best{+.333} & \best{+.355} & \best{+.353} \\
\bottomrule
\end{tabular}
\end{minipage}
\hfill
\begin{minipage}[t]{0.47\textwidth}
\centering
\caption{\textbf{Discrepancy-family ablation at $\tau=0.90$.} Values are mean \texttt{Capture@20} seeds.}
\label{tab:family_ablation}
\begin{tabular}{lcccc}
\toprule
\textbf{Dataset} & \textbf{Learned} & \textbf{Disc.} & \textbf{Stab.}\\
\midrule
Adult & \best{.718} & .188 & .116 \\
Bank & \best{.740} & .289 & .203 \\
MiniBooNE & \best{.669} & .331 & .245 \\
Magic & \best{.345} & .138 & .047 \\
Nomao & .561 & .354 & \best{.728} \\
Spambase & .279 & .287 & \best{.437} \\
Phoneme & \best{.250} & .167 & .167 \\
\bottomrule
\end{tabular}
\vspace{-1.8em}
\end{minipage}

\end{table}

The central empirical question is not whether a new model improves global predictive accuracy, but whether false-confidence concentration is recoverable and regime-consistent with the structural characterization above. We read the evidence along four axes: performance against standard baselines, agreement across concentration-oriented measures, sensitivity to dataset characteristics, and robustness across strong tabular backbones. Adult, Bank Marketing, and MiniBooNE are the strongest regimes: the strongest prior baselines recover only a modest share of false-confidence mass, whereas the discrepancy family surfaces a large majority under the same review budget. The identity of the strongest family member is informative rather than incidental: learned discrepancy is strongest when multiple conflict cues must be combined jointly, while stability-centered ranking is strongest on Nomao and Spambase, where perturbation fragility carries more separating signal. Magic Telescope is mixed because stronger modern tabular predictors absorb part of the relevant local structure into the base predictor, reducing residual headroom for post-hoc discovery. Phoneme is treated as a boundary regime because sparse false-confidence events reduce the evidential strength of both \texttt{FalseConf-AUROC} and \texttt{Capture@20}. Table~\ref{tab:family_ablation} should therefore be read mechanistically: raw confidence ranking is a weak baseline, whereas supervised certainty features reveal that confidence initiates the upper tail in the strongest regimes; support and stability provide the state variables needed for localization, regime transfer, and stability-dominated cases. Appendix Tables~\ref{tab:extended_validation_compact_a}--\ref{tab:extended_validation_compact_b} report the corresponding signal-family, fixed-backbone, perturbation, uncertainty, downstream calibration, region-quality, and stronger-baseline checks. Appendix Sections~\ref{sec:signal_drivers}--\ref{sec:fixed_anchor} further interpret the driver structure in the strongest regimes, the mixed-regime behavior on Magic Telescope, and the falsification-oriented role of the fixed analytic score. Appendix Table~\ref{tab:all_dataset_uncertainty} reports paired seed-level uncertainty for the same dataset--seed records used in Table~\ref{tab:main_discrepancy_family}, and Appendix Table~\ref{tab:null_concentration} (null concentration) compares the observed \texttt{Family Cap@20} values against random 20\% review slices; the anonymous review artifact in Appendix Section~\ref{sec:artifact_availability} maps these empirical objects to released scripts and stored outputs. This null-separation matters because it shows that the main effect is not a generic review-budget artifact: in the strongest regimes, discrepancy-aware ranking recovers far more false-confidence mass than a prevalence-matched random slice, so the reported gains reflect structured localization rather than merely inspecting more cases.

\begin{table}[t]
\centering
\tiny
\setlength{\tabcolsep}{2.2pt}
\renewcommand{\arraystretch}{0.92}

\begin{minipage}[t]{0.49\textwidth}
\centering
\caption{\textbf{Extended validation.} Values at $\tau=0.90$. Panels A--B report \texttt{Cap@20}; Panel C reports $\Delta$\texttt{Cap@20}, except the last row.}
\label{tab:extended_validation_compact_a}
\begin{tabular}{llcccc}
\toprule
\textbf{P} & \textbf{Setting} & \textbf{Adult} & \textbf{Bank} & \textbf{MiniB} & \textbf{Mean} \\
\midrule
\multicolumn{6}{l}{\textit{A. Signal-family necessity}} \\
& Full model & 0.744 & 0.750 & 0.676 & 0.723 \\
& No certainty & 0.303 & 0.419 & 0.383 & 0.368 \\
& No support & 0.774 & 0.813 & 0.698 & 0.761 \\
& No stability & 0.743 & \best{0.827} & 0.782 & 0.784 \\
& Certainty-only & \best{0.780} & 0.822 & \best{0.788} & \best{0.797} \\
& Support-only & 0.272 & 0.360 & 0.364 & 0.332 \\
& Stability-only & 0.192 & 0.326 & 0.341 & 0.286 \\
\midrule
\multicolumn{6}{l}{\textit{B. Fixed-backbone robustness}} \\
& HGB & 0.722 & 0.749 & 0.707 & 0.726 \\
& XGBoost & 0.725 & \best{0.750} & 0.676 & 0.717 \\
& CatBoost & \best{0.736} & 0.748 & \best{0.751} & \best{0.745} \\
\midrule
\multicolumn{6}{l}{\textit{C. Perturbation sensitivity}} \\
& $\lambda=.05$ & 0.640 & 0.464 & 0.172 & 0.425 \\
& $\lambda=.10$ & \best{0.642} & 0.479 & 0.267 & 0.463 \\
& $\lambda=.20$ & 0.624 & 0.528 & 0.382 & 0.511 \\
& $\lambda=.30$ & 0.629 & \best{0.579} & \best{0.417} & \best{0.542} \\
& $k=5$ & 0.411 & 0.406 & 0.269 & 0.362 \\
& $k=10$ & 0.517 & 0.495 & 0.314 & 0.442 \\
& $k=20$ & \best{0.602} & \best{0.534} & \best{0.347} & \best{0.495} \\
& 95\% CI & [.582,.762] & [.393,.612] & [.279,.440] & -- \\
\bottomrule
\end{tabular}
\end{minipage}
\hfill
\begin{minipage}[t]{0.49\textwidth}
\centering
\caption{\textbf{Consequences and baselines.} Calibration, region quality, and stronger comparator checks on the three strongest datasets.}
\label{tab:extended_validation_compact_b}
\begin{tabular}{llcccc}
\toprule
\textbf{P} & \textbf{Analysis} & \textbf{Adult} & \textbf{Bank} & \textbf{MiniB} & \textbf{Mean} \\
\midrule
\multicolumn{6}{l}{\textit{A. Downstream calibration}} \\
& Unweighted ECE & 0.006 & 0.011 & 0.004 & 0.007 \\
& Weighted ECE & \best{0.004} & \best{0.010} & \best{0.003} & \best{0.006} \\
& Unweighted Brier & 0.086 & 0.061 & 0.040 & 0.063 \\
& Weighted Brier & \best{0.065} & \best{0.051} & \best{0.033} & \best{0.058} \\
& Unweighted NLL & 0.273 & 0.195 & 0.136 & 0.202 \\
& Weighted NLL & \best{0.220} & \best{0.157} & \best{0.117} & \best{0.192} \\
& Unweighted Cap@20 & 0.022 & 0.008 & 0.008 & 0.012 \\
& Weighted Cap@20 & \best{0.024} & \best{0.012} & \best{0.009} & \best{0.018} \\
\midrule
\multicolumn{6}{l}{\textit{B. Region quality}} \\
& Top-1 FC mass & \best{0.736} & 0.667 & 0.410 & 0.604 \\
& Top-2 FC mass & \best{0.902} & 0.842 & 0.768 & 0.837 \\
& Top-region support & 0.950 & 0.943 & \best{0.926} & 0.939 \\
& Top-region instability & 0.008 & 0.007 & \best{0.022} & 0.012 \\
\midrule
\multicolumn{6}{l}{\textit{C. Stronger baselines}} \\
& Failure predictor AUROC & 0.483 & 0.655 & 0.708 & 0.615 \\
& Failure predictor Cap@20 & 0.000 & 0.063 & 0.249 & 0.104 \\
& Selective / AURC & 0.028 & 0.014 & 0.007 & 0.016 \\
& Disc. family AUROC & \best{0.850} & \best{0.858} & \best{0.833} & \best{0.847} \\
& Disc. family Cap@20 & \best{0.744} & \best{0.750} & \best{0.676} & \best{0.723} \\
\bottomrule
\end{tabular}
\end{minipage}
\end{table}

\section{Discussion and Limitations}

The principal result is not that one new score dominates every dataset, but that calibration failure can be treated as a discoverable structural object. FALCON-Discover shows that high-confidence errors can concentrate in compact regions of prediction space, that these regions can be recovered under fixed review budgets, and that the recovered regions can be described through a discrepancy state rather than only through a scalar confidence value. This changes the target of calibration analysis: instead of fitting only a global probability map, one can first identify where confidence is misleading and then allocate monitoring, abstention, or calibration effort selectively. The contribution is therefore methodological in a precise sense. The paper does not propose a new end-to-end classifier, but a new post-hoc learning paradigm---prediction-to-structure learning---in which held-out prediction behavior is lifted into a structured conflict state and only then queried through ranking, localization, or calibration-facing actions. The witness scores, learned witness detector, and downstream weighting profile are all operational views of this same state, so the novelty lies not in architectural capacity but in the introduction of a new reliability object together with a reusable state-based interface for acting on it. Although a global calibrator optimized for average probability quality and a discrepancy-aware weighting scheme emphasize different objectives, they are not in conflict in the present benchmark: the weighted stage improves both calibration quality and dangerous-error recovery. Discovery therefore remains the main contribution, while calibration-facing weighting shows that the discovered discrepancy structure is operationally actionable. The study deliberately starts from binary tabular benchmarks, where confidence, support, neighborhood agreement, and support-preserving perturbations can be defined without modality-specific assumptions. This is a controlled identification setting rather than a claim of completed generality: the contribution is the reliability object and prediction-to-structure interface, not a universal perturbation recipe for every modality. Because the strongest discrepancy-family member varies across datasets, FALCON-Discover should be read as a family-level discovery framework rather than a single universally dominant score. Extending the same principle to multiclass, sequential, vision, language, and multimodal systems requires modality-specific definitions of support and stability, and is the natural next test of generality.
\vspace{-0.5em}

\section{Conclusion}

This paper introduced false-confidence concentration as a structural object for calibration research and presented FALCON-Discover as a post-hoc prediction-to-structure framework for making that object measurable, localizable, and actionable. The central finding is that dangerous overconfidence can be concentrated in compact, discoverable subsets of predictions rather than diffusely distributed across the test set. This shifts calibration from global score alignment alone toward targeted reliability discovery: identifying where confidence becomes unsafe to trust. Empirically, concentration is strong in several benchmark regimes, mixed when stronger predictors absorb local structure, and weakly interpretable when false-confidence events become sparse. Methodologically, the results separate raw confidence ranking from learned certainty structure and from the broader discrepancy state: confidence can initiate the upper tail, while support and stability make the failure region interpretable, transferable across regimes, and actionable for calibration-facing weighting. The strongest claim is therefore family-level: reliability failure is structured, not merely scalar. Calibration should not only align scores with frequencies, but also reveal where predictions should not be trusted.

\bibliographystyle{plain}
\bibliography{references}

\clearpage
\appendix
\onecolumn

\clearpage

\section{Evaluation Reporting Details}
\label{sec:eval_reporting_details}

The evidence is organized around four aspects: comparison with standard aggregate calibration and trust-scoring baselines, comparison across complementary concentration-oriented views, sensitivity to data characteristics, and sensitivity to classifier characteristics. Concretely, the empirical evidence is reported through six quantities: FalseConf-AUROC, Capture@20, dataset-level operating-point summaries, absolute event-count summaries on the strongest datasets, threshold robustness under $\tau\in\{0.85,0.90,0.95\}$, and benchmark-level synthesis across datasets and detector variants, rather than raw accuracy alone. Standard aggregate calibration measures such as ECE, Brier score, NLL, and reliability diagrams remain important reference quantities, but they do not directly answer the paper’s operational question of whether dangerous high-confidence errors are concentrated and recoverable within a small review slice. This combination is deliberate: it ensures that the claims do not rely on one metric family alone and that the core empirical story is supported simultaneously by ranking quality, operational impact, visual evidence, robustness analysis, sensitivity to event prevalence and class balance, and detector-family ablation across strong tabular backbones. Because the contribution is empirical discovery rather than universal score dominance, the success criterion is precise: the method succeeds if discrepancy-family ranking captures substantially more false-confidence mass than confidence-only and the strongest validation-selected prior baseline, ideally with large effect sizes, stable threshold behavior, and interpretable discrepancy regions.

\section{Additional Visual Evidence}

This section provides representative operating-point summaries and discrepancy-region visualizations for the strongest datasets, complementing the quantitative results in the main text. These figures are included as diagnostic visual evidence rather than primary quantitative evidence; the main claims are supported by the tables, bootstrap intervals, threshold sweeps, and backbone checks.

\begin{figure}[H]
    \centering
    \begin{subfigure}{0.48\textwidth}
        \centering
        \includegraphics[width=\linewidth]{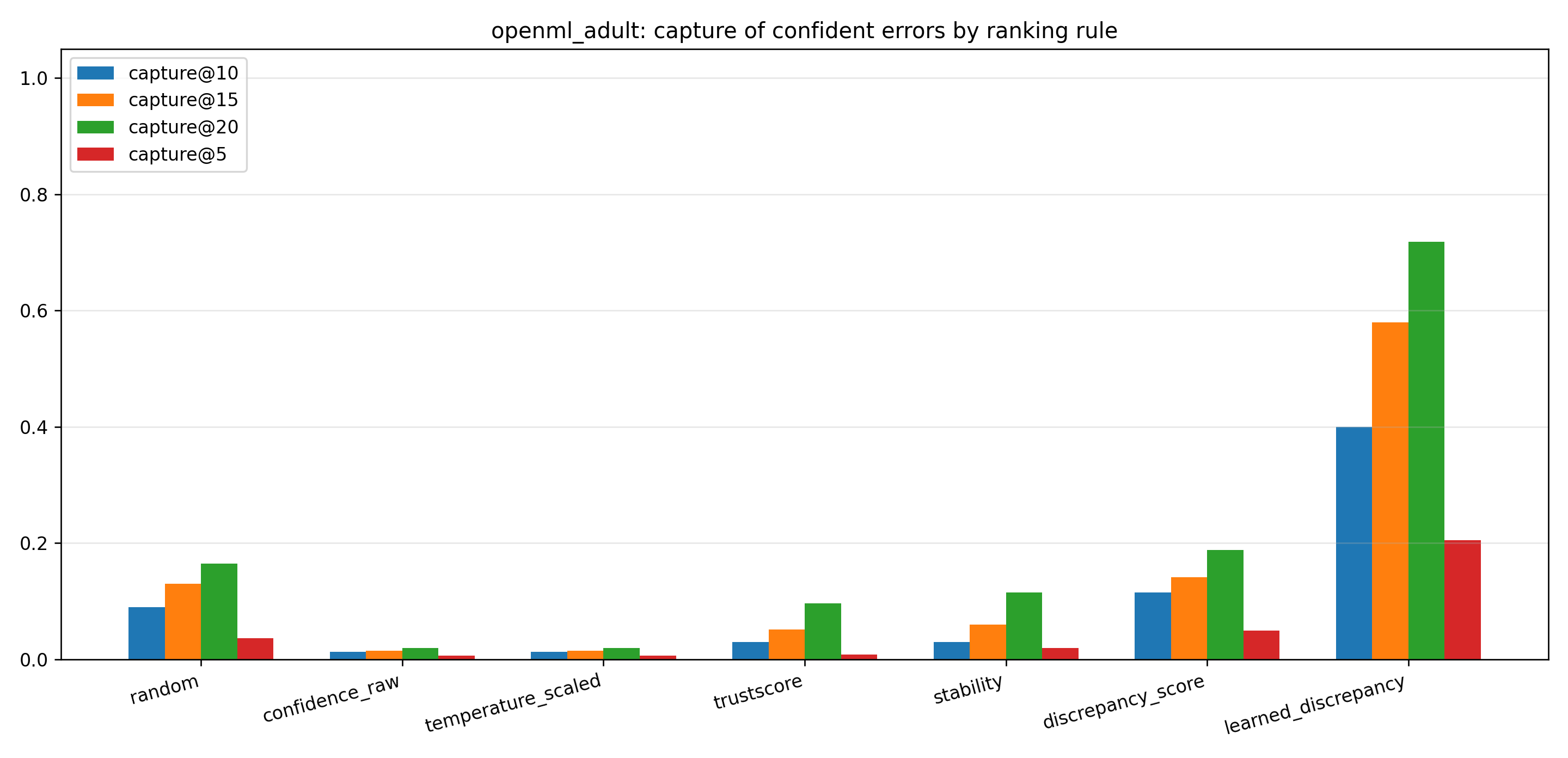}
        \caption{Adult: operating-point summary}
    \end{subfigure}
    \hfill
    \begin{subfigure}{0.48\textwidth}
        \centering
        \includegraphics[width=\linewidth]{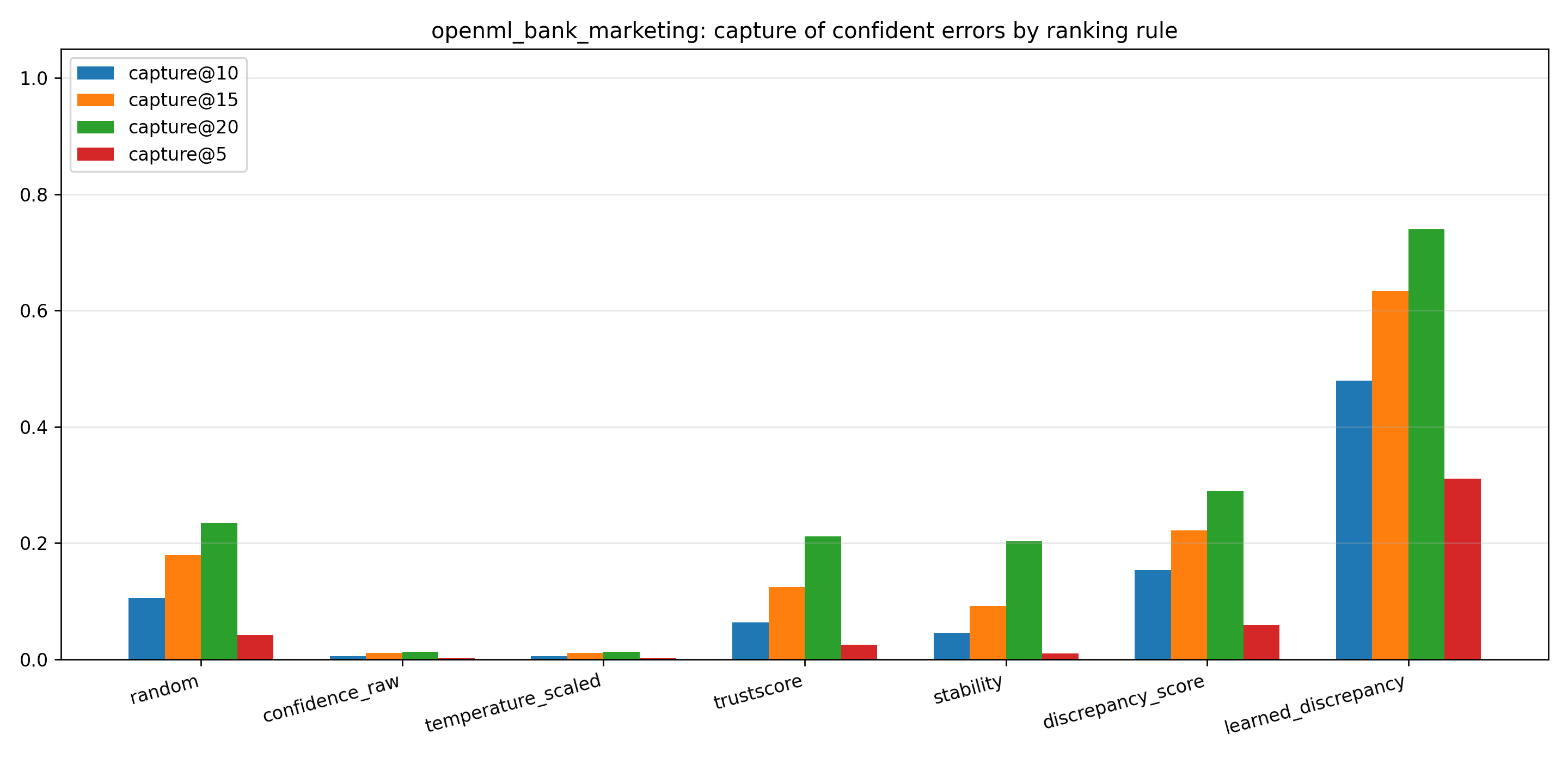}
        \caption{Bank Marketing: operating-point summary}
    \end{subfigure}

    \vspace{0.4em}

    \begin{subfigure}{0.48\textwidth}
        \centering
        \includegraphics[width=\linewidth]{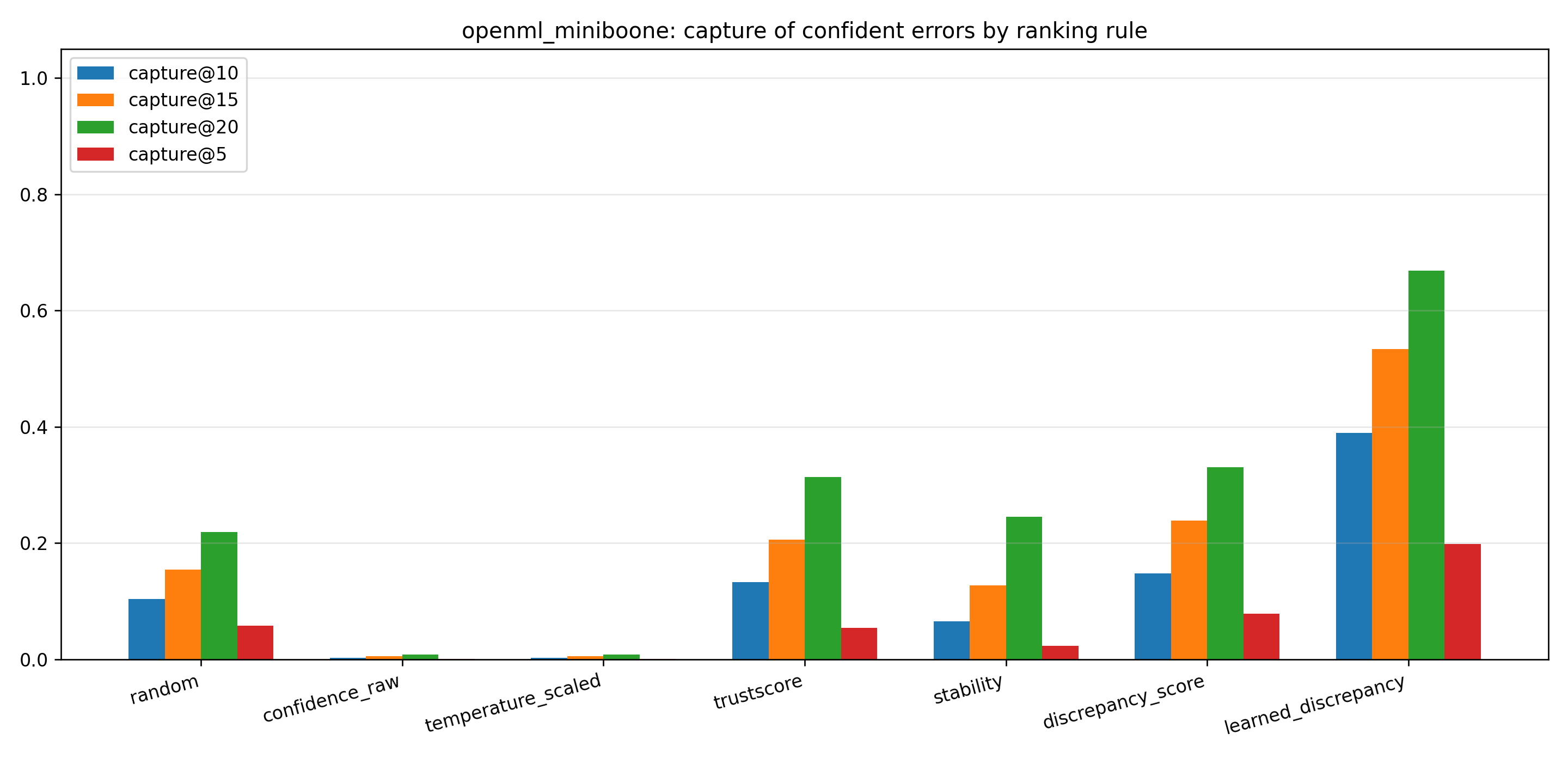}
        \caption{MiniBooNE: operating-point summary}
    \end{subfigure}
    \hfill
    \begin{subfigure}{0.48\textwidth}
        \centering
        \includegraphics[width=\linewidth]{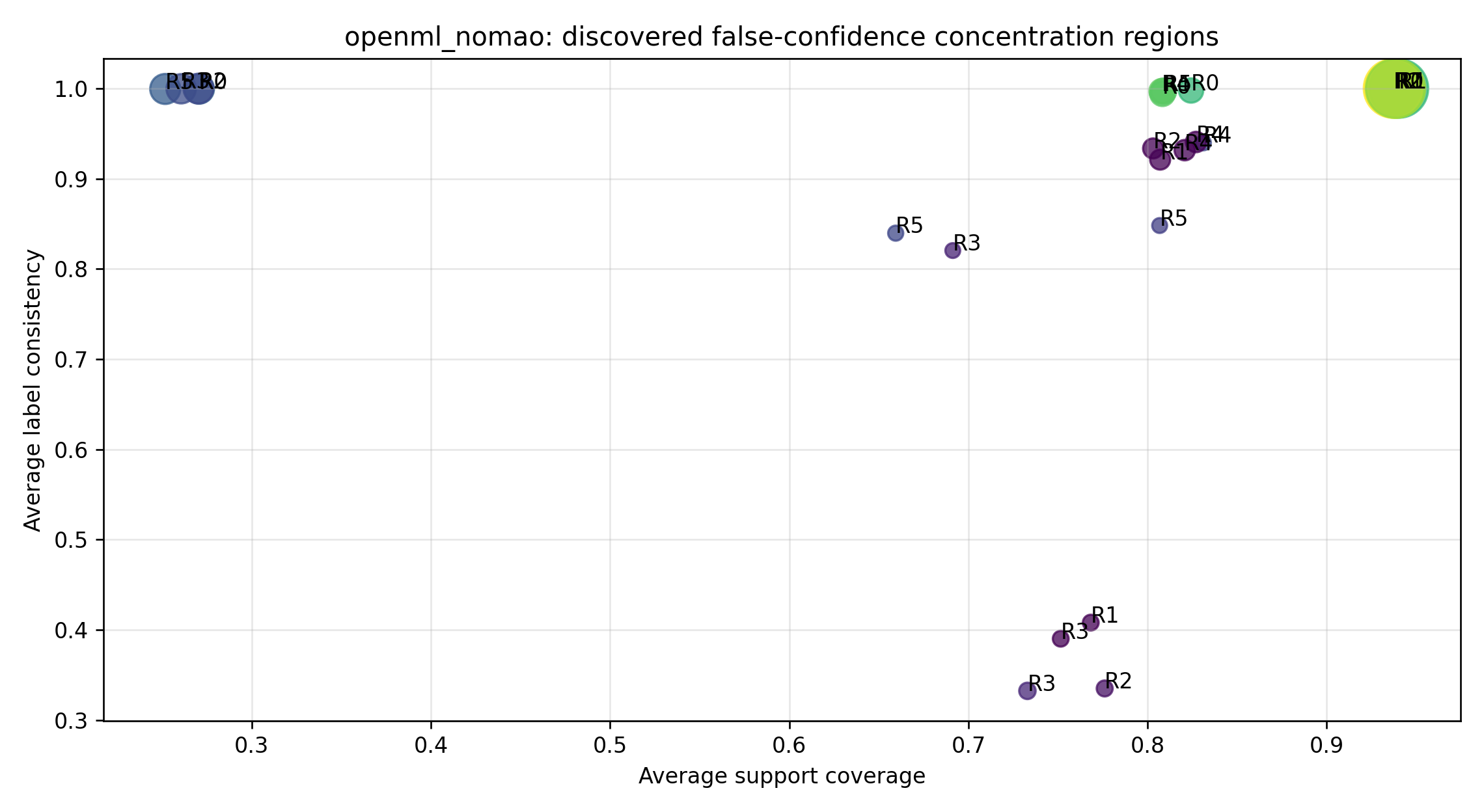}
        \caption{Nomao: discrepancy-region structure}
    \end{subfigure}
    \caption{\textbf{Additional visual evidence.} Discrepancy-family ranking recovers substantially more high-confidence errors than prior baselines on the strongest datasets, while discrepancy regions are structurally organized rather than diffuse.}
    \label{fig:main_visuals}
\end{figure}

\clearpage
\section{Per-dataset operating-point summaries}

This appendix contains the full per-dataset operating-point summaries, the full per-dataset discrepancy-region visualizations, and the full per-dataset discrepancy-aware weighting profiles.

\begin{figure}[H]
    \centering
    \begin{subfigure}{0.48\textwidth}
        \centering
        \includegraphics[width=\linewidth]{figures/openml_adult/fig1_global_metrics.png}
        \caption{Adult}
    \end{subfigure}
    \hfill
    \begin{subfigure}{0.48\textwidth}
        \centering
        \includegraphics[width=\linewidth]{figures/openml_bank_marketing/fig1_global_metrics.png}
        \caption{Bank Marketing}
    \end{subfigure}

    \vspace{0.4em}

    \begin{subfigure}{0.48\textwidth}
        \centering
        \includegraphics[width=\linewidth]{figures/openml_miniboone/fig1_global_metrics.png}
        \caption{MiniBooNE}
    \end{subfigure}
    \hfill
    \begin{subfigure}{0.48\textwidth}
        \centering
        \includegraphics[width=\linewidth]{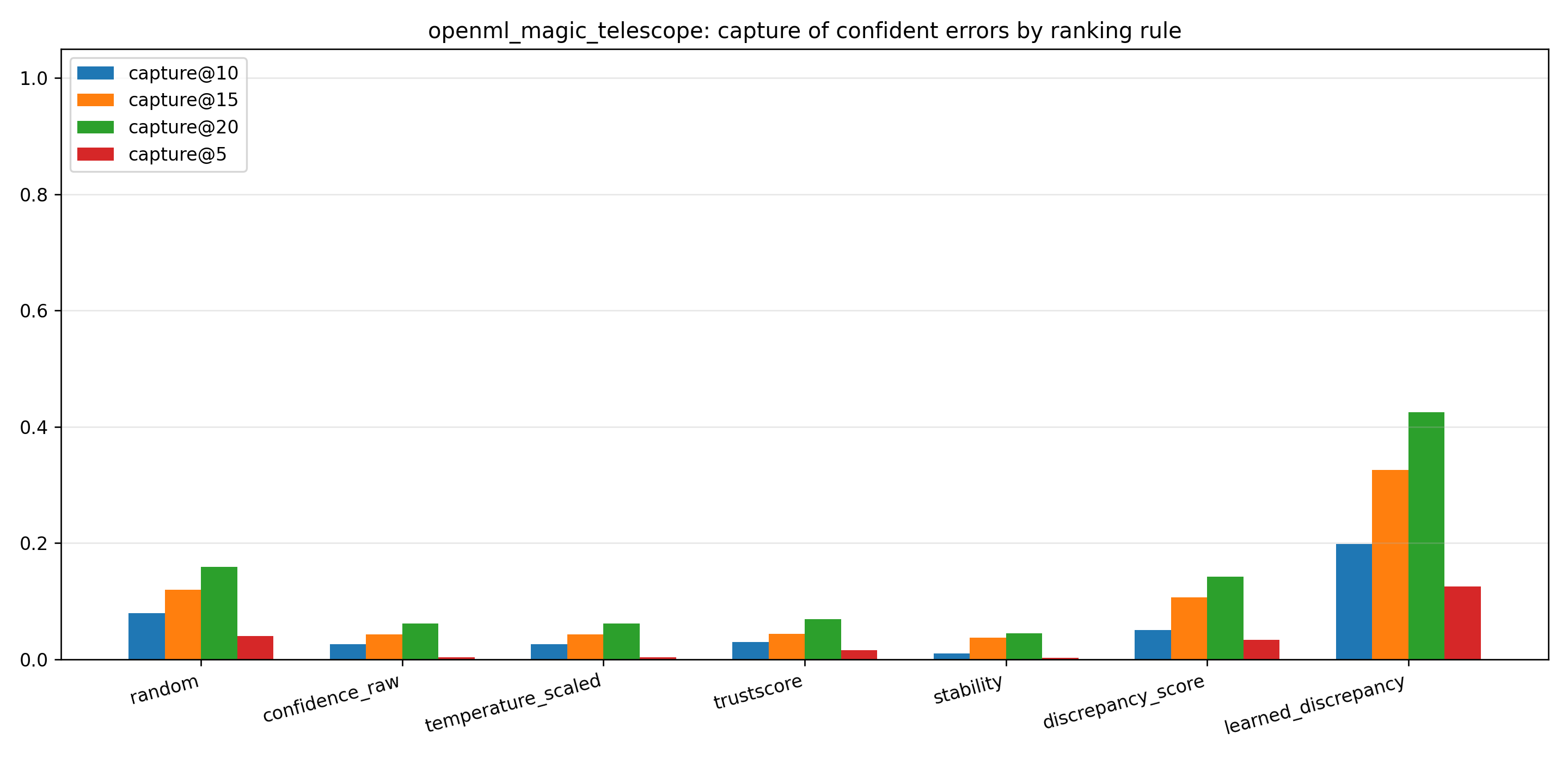}
        \caption{Magic Telescope}
    \end{subfigure}
\end{figure}

\begin{figure}[H]
    \centering
    \begin{subfigure}{0.48\textwidth}
        \centering
        \includegraphics[width=\linewidth]{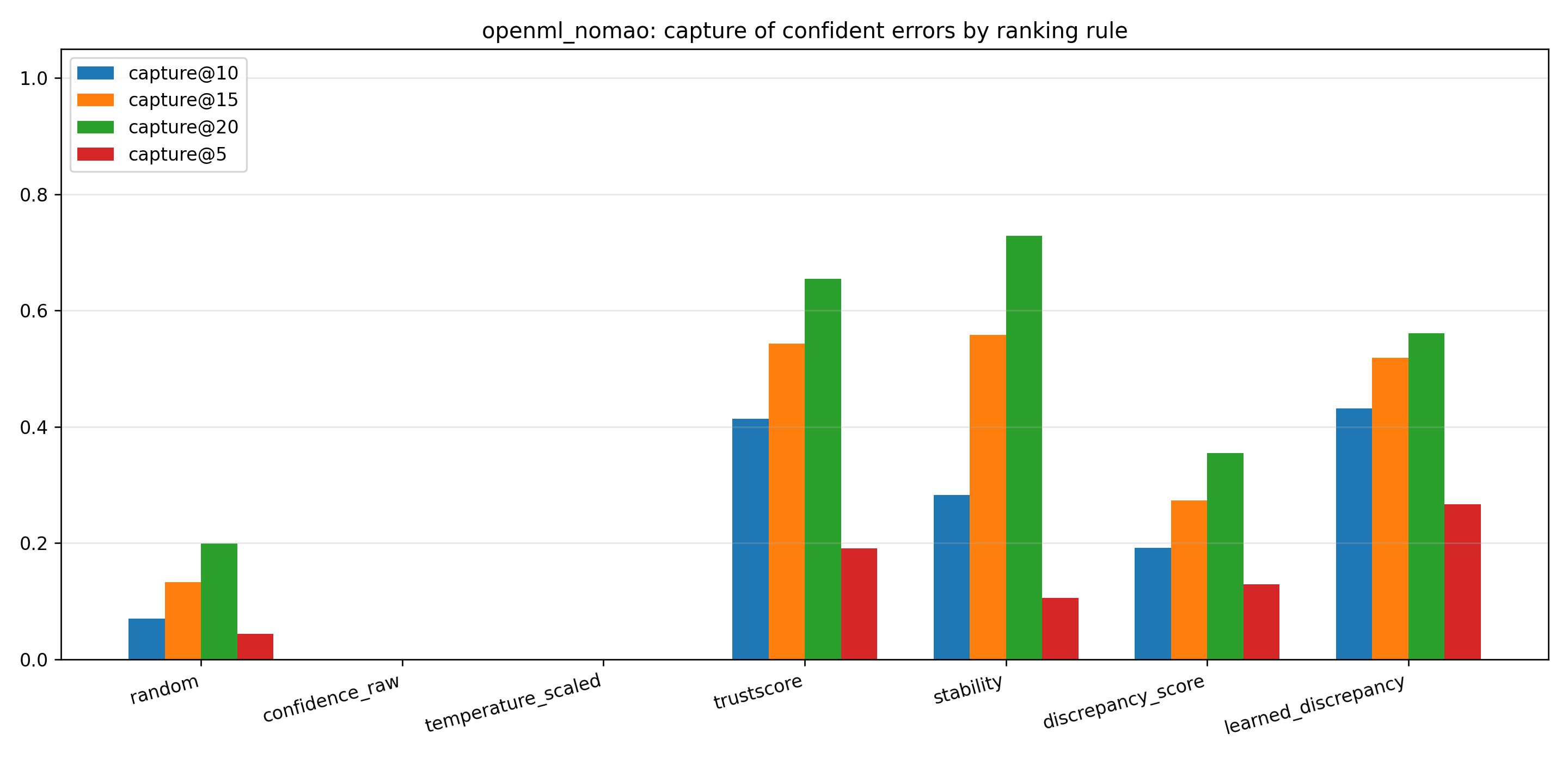}
        \caption{Nomao}
    \end{subfigure}
    \hfill
    \begin{subfigure}{0.48\textwidth}
        \centering
        \includegraphics[width=\linewidth]{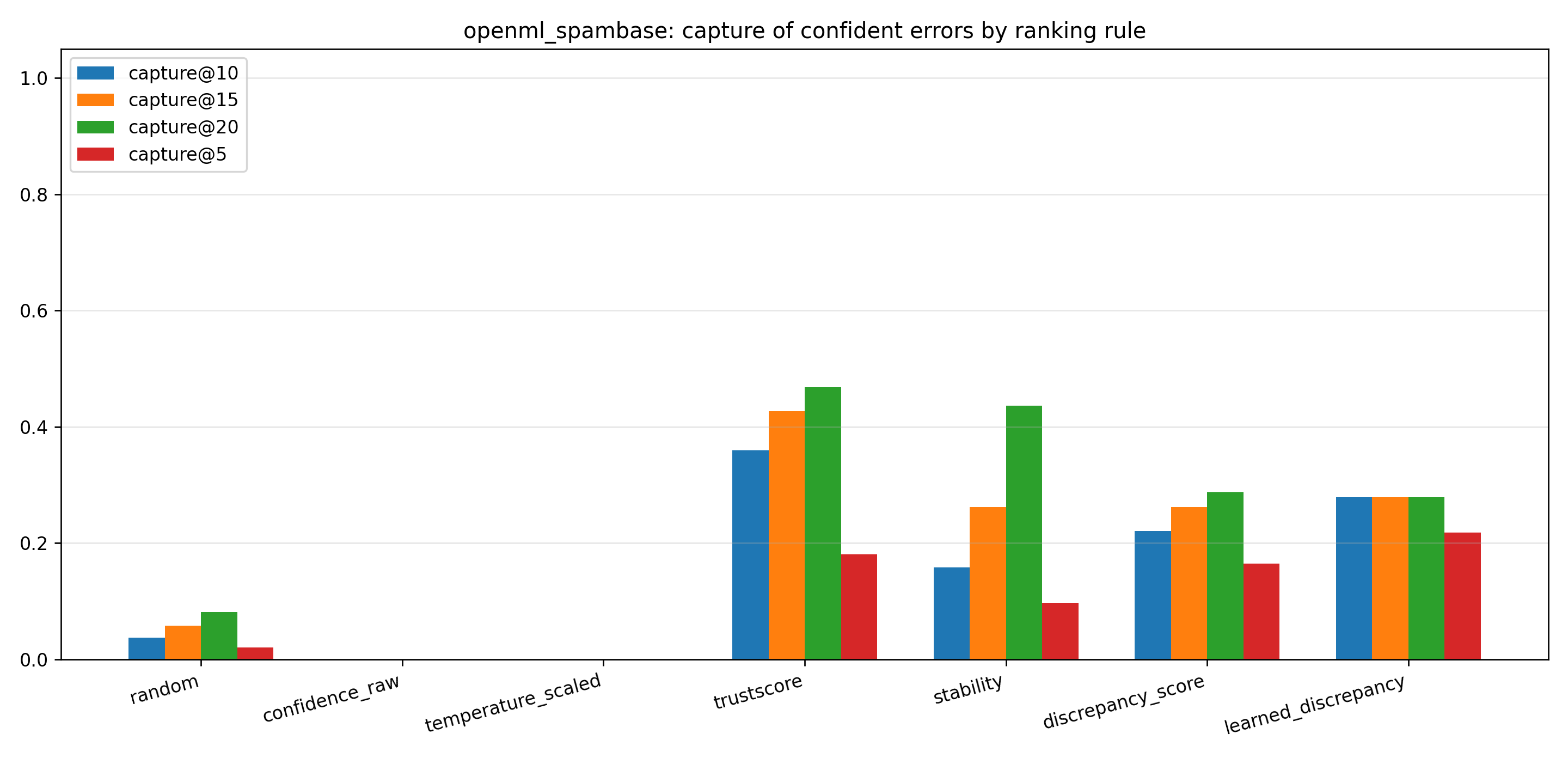}
        \caption{Spambase}
    \end{subfigure}

    \vspace{0.4em}

    \begin{subfigure}{0.48\textwidth}
        \centering
        \includegraphics[width=\linewidth]{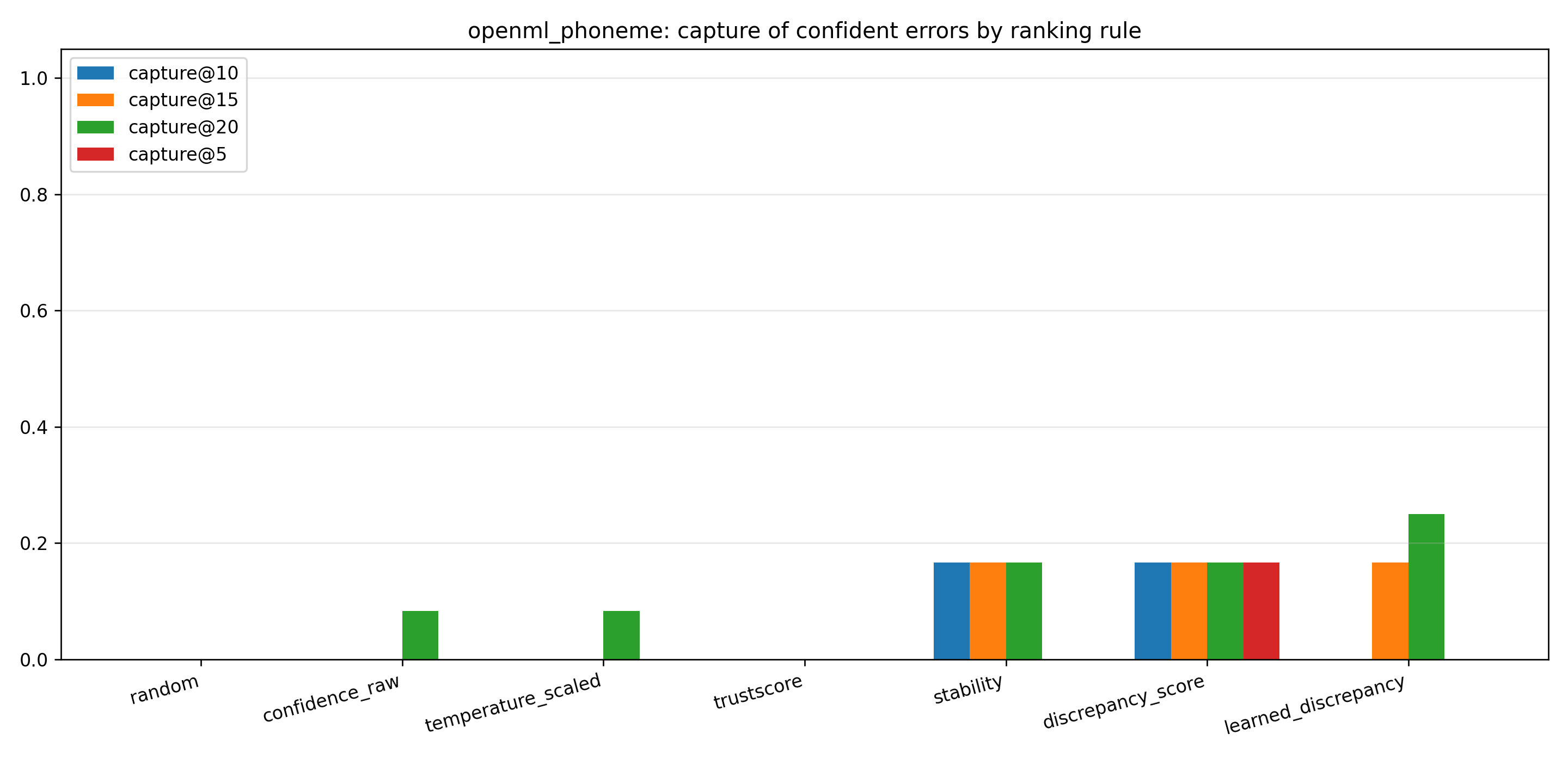}
        \caption{Phoneme}
    \end{subfigure}
    \caption{\textbf{Full per-dataset operating-point summaries available in the anonymous review artifact.}}
\end{figure}

\clearpage
\section{Per-dataset discrepancy-region visualizations}

\begin{figure}[H]
    \centering
    \begin{subfigure}{0.48\textwidth}
        \centering
        \includegraphics[width=\linewidth]{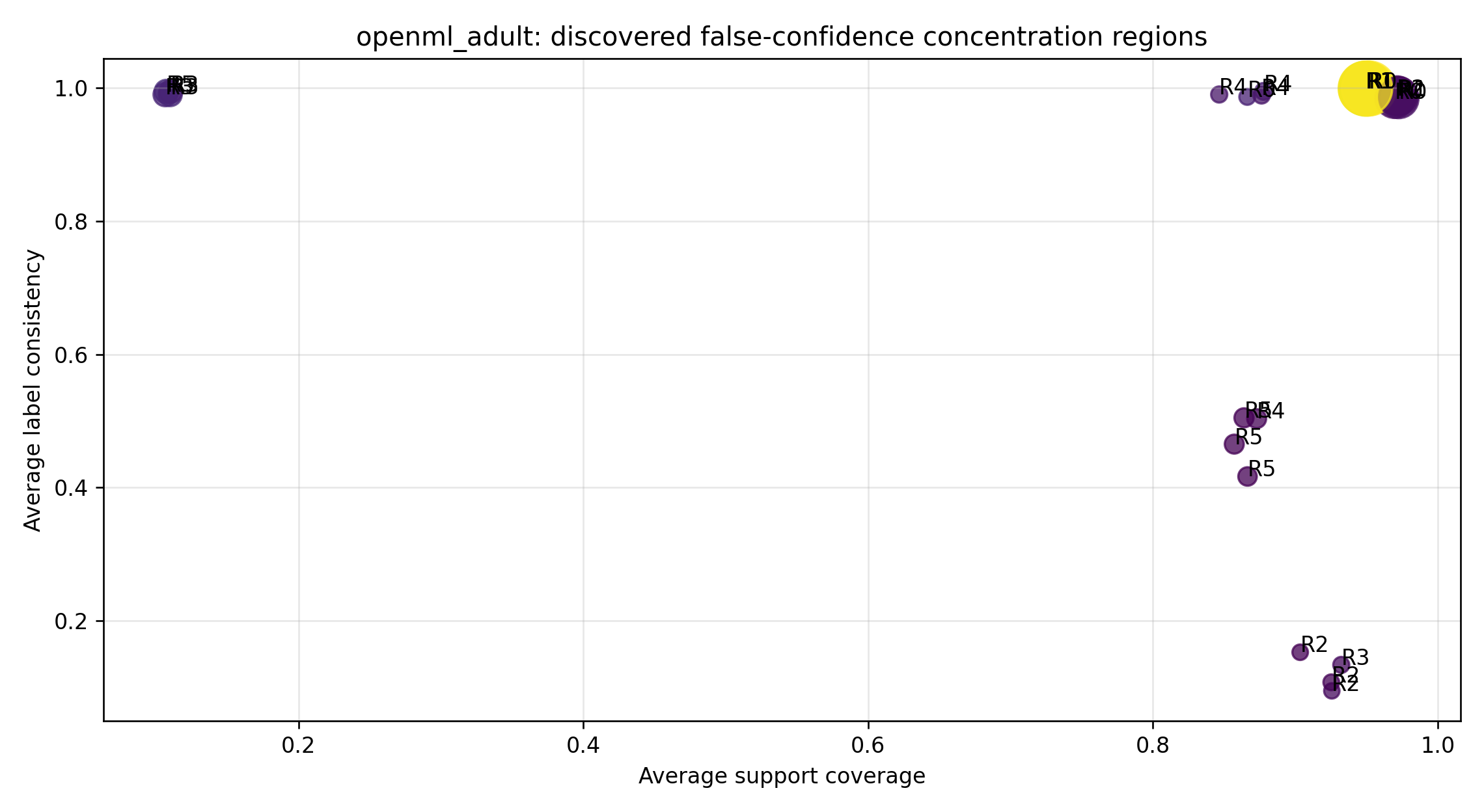}
        \caption{Adult}
    \end{subfigure}
    \hfill
    \begin{subfigure}{0.48\textwidth}
        \centering
        \includegraphics[width=\linewidth]{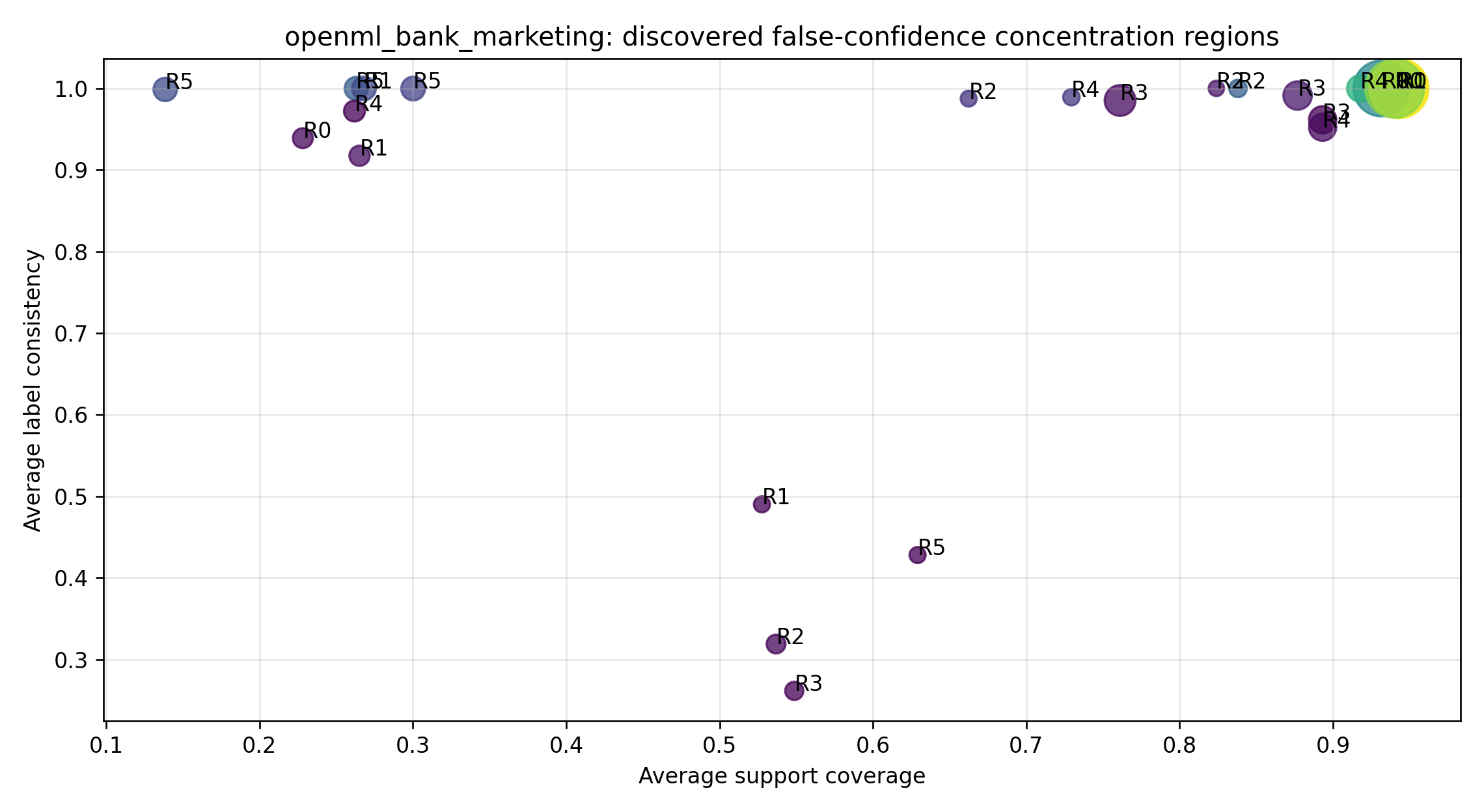}
        \caption{Bank Marketing}
    \end{subfigure}

    \vspace{0.4em}

    \begin{subfigure}{0.48\textwidth}
        \centering
        \includegraphics[width=\linewidth]{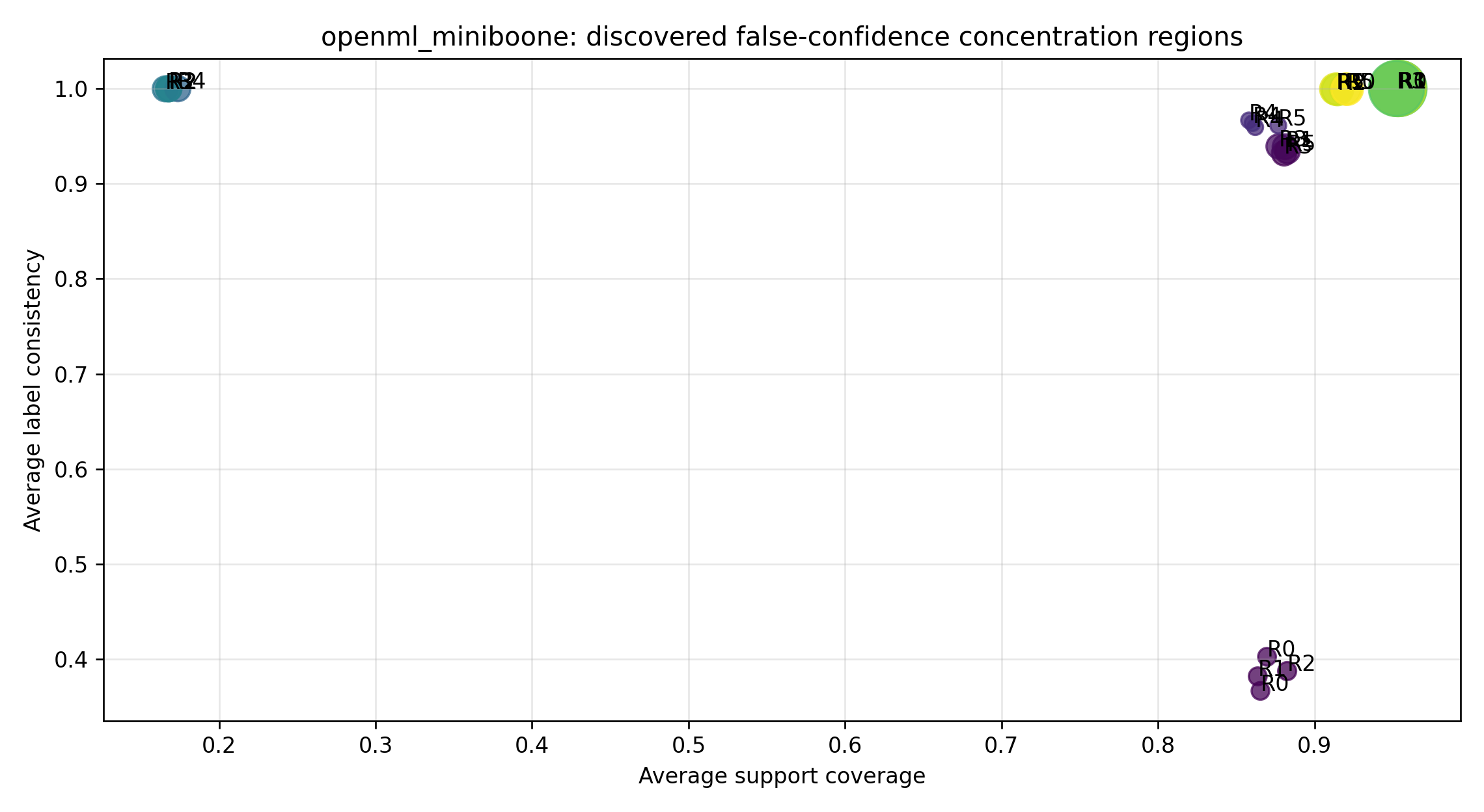}
        \caption{MiniBooNE}
    \end{subfigure}
    \hfill
    \begin{subfigure}{0.48\textwidth}
        \centering
        \includegraphics[width=\linewidth]{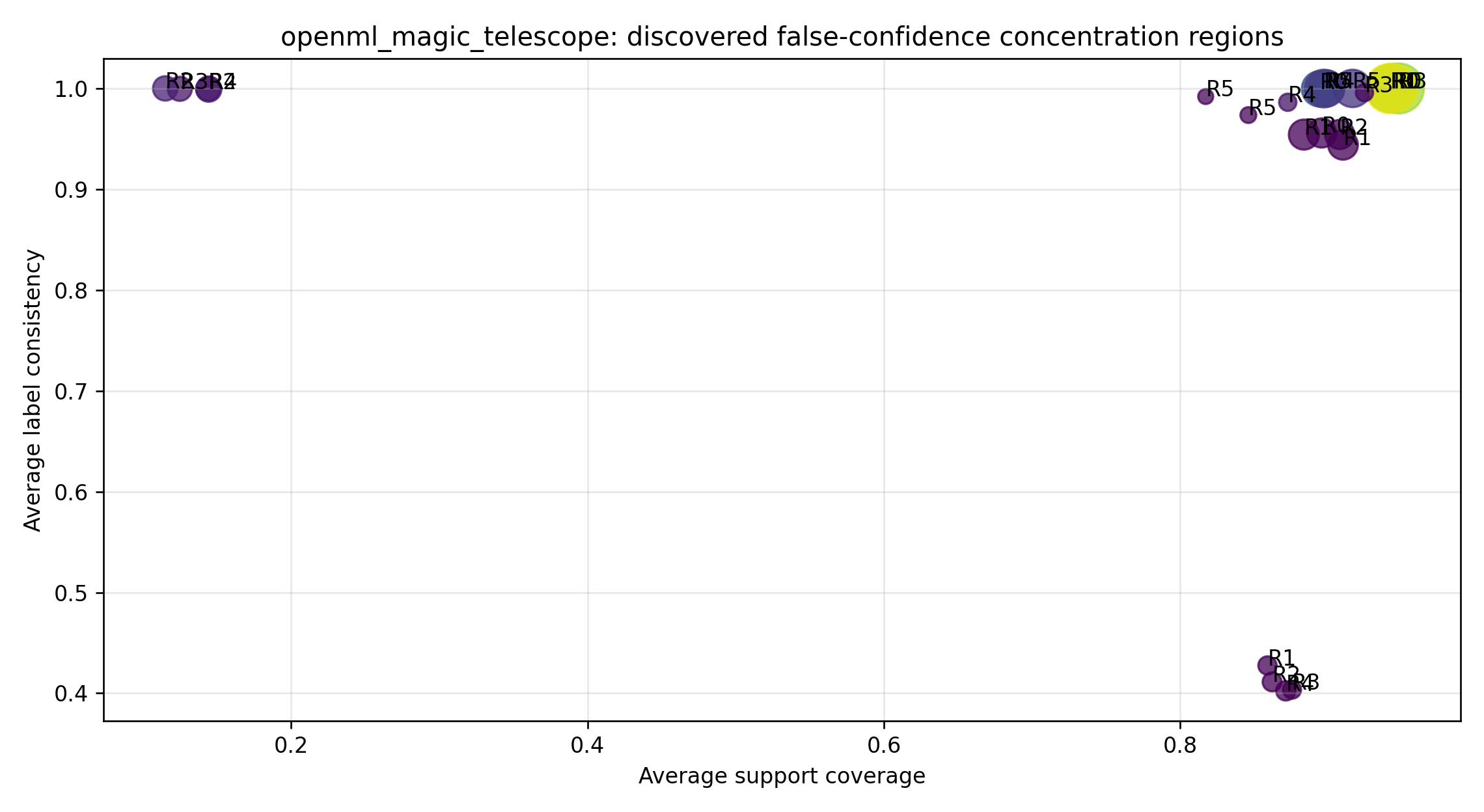}
        \caption{Magic Telescope}
    \end{subfigure}
\end{figure}

\begin{figure}[H]
    \centering
    \begin{subfigure}{0.48\textwidth}
        \centering
        \includegraphics[width=\linewidth]{figures/openml_nomao/fig2_discrepancy_clusters.png}
        \caption{Nomao}
    \end{subfigure}
    \hfill
    \begin{subfigure}{0.48\textwidth}
        \centering
        \includegraphics[width=\linewidth]{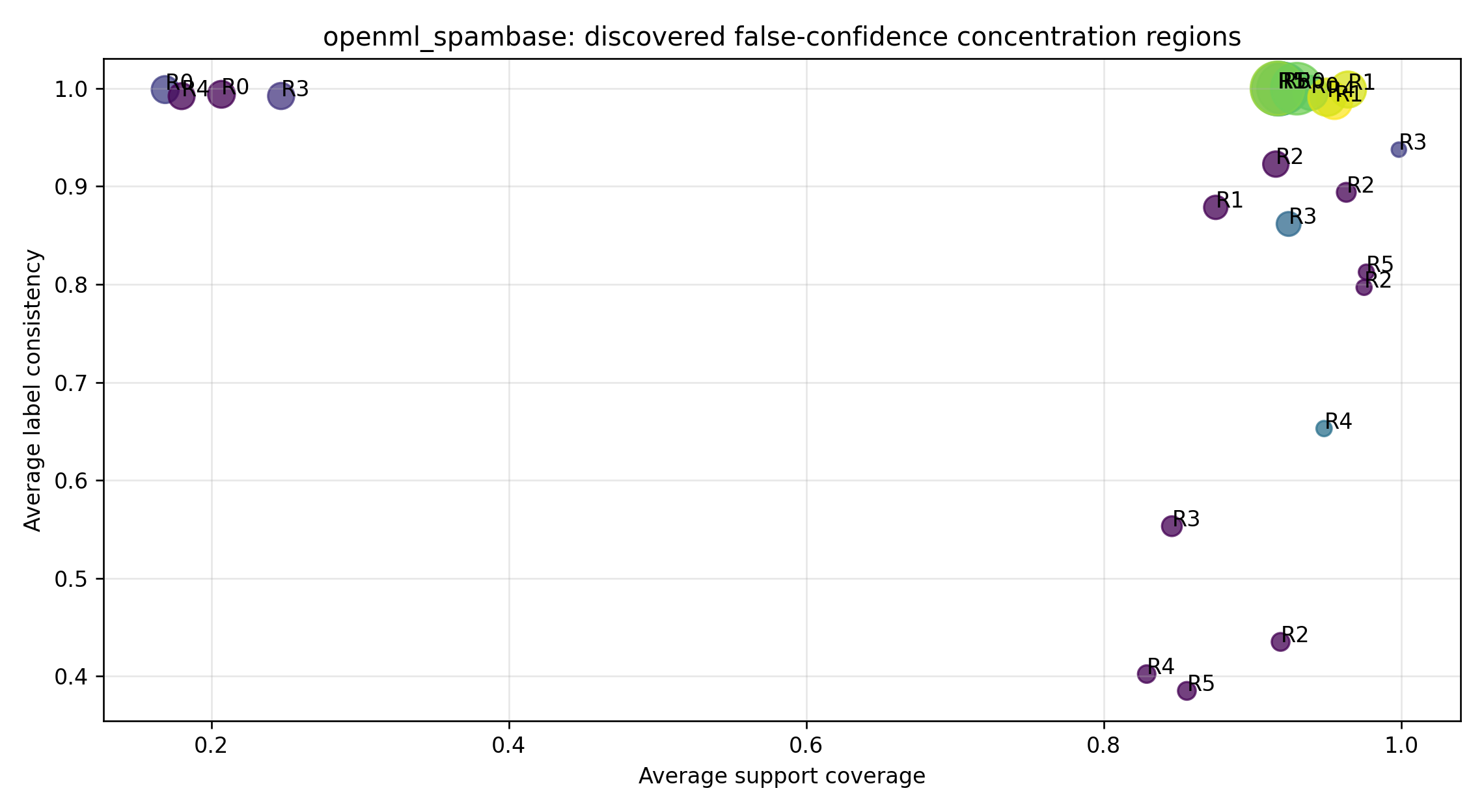}
        \caption{Spambase}
    \end{subfigure}

    \vspace{0.4em}

    \begin{subfigure}{0.48\textwidth}
        \centering
        \includegraphics[width=\linewidth]{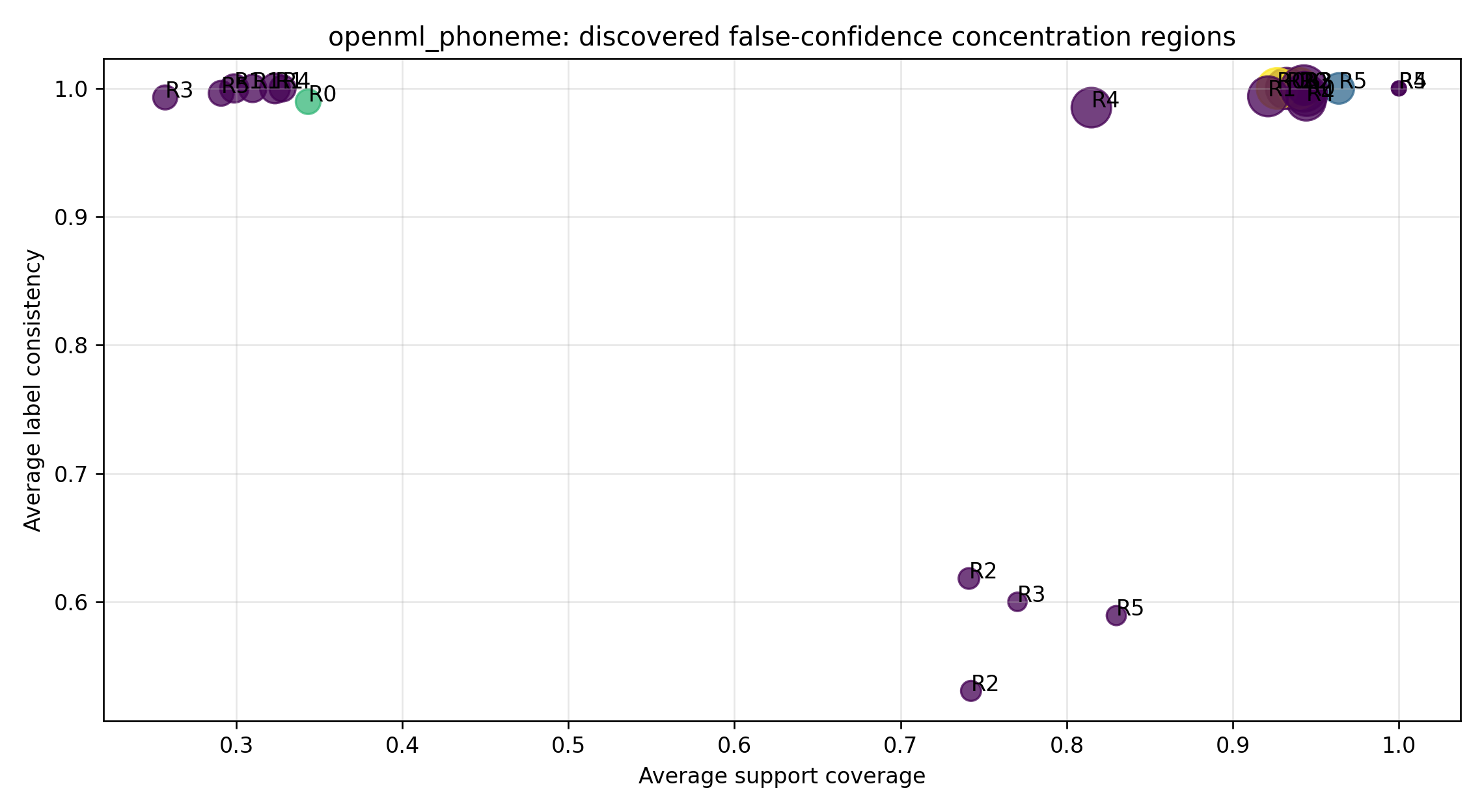}
        \caption{Phoneme}
    \end{subfigure}
    \caption{\textbf{Full per-dataset discrepancy-region visualizations available in the anonymous review artifact.}}
\end{figure}

\clearpage
\section{Interpreting the Signal-Family Ablation}
\label{sec:signal_family_interpretation}

Table~\ref{tab:extended_validation_compact_a} is intended as a mechanistic analysis rather than a claim that every signal family is equally necessary on every dataset. The ablation separates three distinct objects that are easy to conflate. First, \emph{raw confidence ranking} is the minimal deployment baseline: it sorts predictions directly by native confidence and asks whether the model's own confidence is sufficient to expose dangerous errors. This baseline is weak in the strongest regimes, showing that the model's native confidence ordering does not by itself recover the false-confidence mass efficiently. Second, \emph{certainty-only learning} is not the same object as raw confidence ranking. It is a supervised witness detector trained on calibration folds to recognize the false-confidence event using certainty-derived features. Its strength in Adult, Bank Marketing, and MiniBooNE shows that false-confidence concentration often begins in the high-certainty tail. This supports, rather than contradicts, the concentration hypothesis: the dangerous region is not uniformly spread across the prediction space. Third, the \emph{full discrepancy state} is the paper's claim-bearing object. Its purpose is not only to maximize one Capture@20 number on the strongest datasets, but to provide a reusable state for ranking, localization, regime interpretation, and calibration-facing weighting. Support and stability are therefore evaluated by their role in explaining and transporting the discovered failure structure, not only by whether their removal lowers Capture@20 in every certainty-dominated regime. This distinction is important because the strongest signal family changes across regimes: certainty dominates the strongest datasets, whereas perturbation stability is decisive in Nomao and Spambase. The ablation should therefore be read as follows. Certainty identifies where the upper tail begins; support and agreement indicate whether that tail is locally grounded; stability indicates whether the decision is fragile under support-preserving perturbations. FALCON-Discover combines these views because false-confidence concentration is a family-level reliability phenomenon, not a universal single-score mechanism.

\section{Split Discipline and Label Access}
\label{sec:split_discipline}

All learned and weighted components follow a strict split discipline. The base predictor is selected using out-of-fold label AUROC on the training partition, then refit on the full training partition and evaluated on validation and test partitions. Discrepancy features for held-out examples are computed from held-out predictions and training-derived support statistics. The learned discrepancy ranker is fitted only on validation/calibration folds where labels are available. Its target is $z_i=\mathrm{FC}_{\tau}(x_i)$ on those folds. The held-out test labels are never used to fit the ranker, choose the ranker, tune the analytic score, set the perturbation parameters, select the confidence threshold, or construct $w_{\mathrm{cal}}$. Test labels are used only once: to compute final reporting quantities such as \texttt{FalseConf-AUROC}, \texttt{Capture@20}, and recovered false-confidence counts. The weighting profile in Equation~\eqref{eq:cal_weight} is therefore a calibration-training object, not a test-time scoring rule requiring unknown labels. In a deployment pipeline, $\mathrm{FC}_{\tau}(x)$ is available only for calibration data with observed outcomes; for unlabeled future examples, the deployable quantities are the label-free components of $\psi(x)$ and the fitted ranker learned from calibration data.

\section{Distinction from Failure Prediction, Selective Classification, and Trust Scoring}
\label{sec:distinction_failure_prediction}

FALCON-Discover differs from standard failure prediction, selective classification, and trust scoring in its target object and output interface. Failure prediction usually asks whether an auxiliary model can predict correctness. Selective classification asks whether a model should accept, reject, or defer a prediction under a desired coverage-risk tradeoff. Trust scoring asks whether an alternative scalar score better reflects local support than native confidence. These are important but score-centric formulations. FALCON-Discover instead asks whether high-confidence errors occupy a compact, recoverable region of prediction space. The output is therefore not only a scalar trust score or a reject decision. The method constructs a discrepancy state that can be queried in multiple ways: ranking false-confidence events under a review budget, localizing discrepancy-heavy regions, interpreting which signal family drives the failure mode, and deriving calibration-facing weights. This makes false-confidence concentration a structural discovery problem rather than only a correctness-prediction problem. This distinction is also reflected in the evaluation. The main outcome is not global accuracy, ECE alone, or selective risk alone, but the recoverability of dangerous high-confidence error mass under fixed review budgets. \texttt{Capture@20} and \texttt{FalseConf-AUROC} therefore measure whether the false-confidence event is concentrated and discoverable, not whether a model is globally better calibrated on average.

\section{Why Binary Tabular Benchmarks Are the First Testbed}
\label{sec:why_binary_tabular}

The binary tabular setting is used as a controlled first testbed because all components of the discrepancy state can be defined without modality-specific assumptions. Confidence is unambiguous, local support can be measured in transformed feature space, neighborhood agreement can be computed directly, and support-preserving perturbations can be implemented as local neighbor mixing. This makes the setting appropriate for testing the core structural claim: whether dangerous high-confidence errors are concentrated and recoverable. The paper does not claim that the same perturbation operator should transfer unchanged to vision, language, sequential, or multimodal settings. In those settings, support and stability must be redefined using modality-appropriate neighborhoods and perturbations. For example, vision systems may require augmentation-preserving neighborhoods, language systems may require semantic-preserving paraphrase neighborhoods, and sequential systems may require temporally constrained perturbations. The general claim is therefore not tied to tabular interpolation; it is that confidence should be evaluated together with local evidence and stability to discover concentrated false-confidence regions.

\section{Sparse False-Confidence Events and Boundary Regimes}
\label{sec:sparse_boundary}

False-confidence concentration metrics become less stable when the number of false-confidence events is very small. In such cases, both \texttt{FalseConf-AUROC} and \texttt{Capture@20} have reduced evidential strength: a small change in the identity of a few high-confidence errors can noticeably alter the measured recovery fraction. For this reason, Phoneme is treated as a boundary regime rather than as decisive counterevidence. This interpretation is stricter than simply excluding the dataset. The dataset remains reported in the main table, but its role is diagnostic: it shows where the proposed concentration view becomes weakly identifiable because the event space itself is sparse. The paper's main empirical claim is therefore not universal dominance across every dataset, but a regime-resolved finding: false-confidence concentration is strong and actionable when the event space is sufficiently populated, mixed when the base predictor absorbs part of the local structure, and boundary-limited when too few false-confidence events remain to support stable discovery.

\section{Perturbation and Neighborhood Sensitivity}
\label{sec:perturbation_sensitivity_interpretation}

The perturbation and neighborhood parameters are fixed before test evaluation and are not selected on the test split. Their role is to define a local stability probe rather than to optimize final performance. The sensitivity results show that the method is not dependent on a single pathological perturbation setting: performance changes smoothly as $\lambda$ and $k$ vary, with stronger recovery when perturbations are large enough to expose local instability but still small enough to remain support-preserving. The important object is therefore not a uniquely optimal value of $\lambda$ or $k$, but the recurrence of the same qualitative pattern: local stability information provides an independent view of false-confidence structure, and this view becomes decisive in stability-dominated regimes. The fixed settings used in the main experiments are chosen to balance locality and sensitivity, while the appendix reports the broader sweep to make this design choice auditable.

\clearpage

\section{Benchmark-level empirical synthesis}

Table~\ref{tab:appendix_benchmark_synthesis} aggregates the strongest reviewer-oriented empirical checks derived from Tables~\ref{tab:main_discrepancy_family}--\ref{tab:family_ablation}. It answers three questions directly. First, how often does the discrepancy family beat the strongest prior baseline? Second, how stable are the gains across thresholds on the strongest datasets? Third, does the overall story depend on a single detector, or does it hold at the family level? The table shows that the paper’s empirical claim is supported simultaneously at the dataset, threshold, and detector-selection levels.

\begin{table}[H]
\centering
\small
\setlength{\tabcolsep}{6pt}
\renewcommand{\arraystretch}{1.10}
\caption{\textbf{Benchmark-level empirical synthesis derived from the main result tables.}}
\label{tab:appendix_benchmark_synthesis}
\begin{tabular}{lc}
\toprule
\textbf{Benchmark-level check} & \textbf{Value} \\
\midrule
Positive $\Delta$FalseConf-AUROC vs strongest prior baseline & 6/7 datasets \\
Positive $\Delta$Capture@20 vs strongest prior baseline & 5/7 datasets \\
Positive $\Delta$Capture@20 on strongest three datasets across $\tau\in\{0.85,0.90,0.95\}$ & 9/9 settings \\
Mean $\Delta$FalseConf-AUROC across all seven datasets & +0.126 \\
Mean $\Delta$Capture@20 across all seven datasets & +0.216 \\
Mean $\Delta$Capture@20 on strongest three datasets at $\tau=0.85$ & +0.480 \\
Mean $\Delta$Capture@20 on strongest three datasets at $\tau=0.90$ & +0.501 \\
Mean $\Delta$Capture@20 on strongest three datasets at $\tau=0.95$ & +0.434 \\
Learned discrepancy selected as best family member & 5/7 datasets \\
Stability selected as best family member & 2/7 datasets \\
\bottomrule
\end{tabular}
\end{table}

Two summary points are especially important. First, the discrepancy family does not rely on a single unusually favorable dataset: it achieves positive AUROC gains on six of seven datasets and positive \texttt{Capture@20} gains on five of seven datasets. Second, the strongest regimes are stable under threshold variation: all nine threshold settings for Adult, Bank Marketing, and MiniBooNE remain positive. This makes the empirical case materially stronger than a one-threshold, one-detector, or one-dataset story.

\clearpage
\section{Dataset-Regime Interpretation}

This section provides a concise summary of how the main evaluation metrics should be interpreted across different empirical regimes in the benchmark. The regimes reflect differences in false-confidence event prevalence, class balance, and structural complexity. This table is referenced in the main paper to guide the interpretation of \texttt{FalseConf-AUROC} and \texttt{Capture@20} under varying data conditions.

\begin{table}[H]
\centering
\tiny
\setlength{\tabcolsep}{5pt}
\renewcommand{\arraystretch}{1.08}
\caption{\textbf{Dataset-regime interpretation.} Summary of how the main evaluation quantities should be read across event prevalence, class balance, and sparse-regime boundary cases.}
\label{tab:dataset_regimes}
\begin{tabular}{lll}
\toprule
\textbf{Regime} & \textbf{Datasets} & \textbf{Interpretation} \\
\midrule
Strong & Adult, Bank Marketing, MiniBooNE &
Both \texttt{FalseConf-AUROC} and \texttt{Capture@20} are highly informative; \\
& & absolute recovery and threshold robustness are also meaningful. \\
Mixed & Magic Telescope, Nomao, Spambase &
Metrics remain useful, but gains are smaller or detector-dependent; \\
& & AUROC and \texttt{Capture@20} should be read jointly. \\
Boundary & Phoneme &
Event sparsity limits interpretability; both metrics should be read cautiously \\
& & as boundary-case evidence rather than decisive negative evidence. \\
\bottomrule
\end{tabular}
\end{table}

\section{Functional Stability of Discrepancy Regions}
\label{sec:functional_region_stability}

A natural question is whether the discovered discrepancy regions are stable across seeds. For this paper, the relevant notion of stability is \emph{functional} rather than label-identitarian. Because unsupervised cluster labels are permutation-invariant, raw cluster identities are not meaningful objects to compare across random seeds. The stable scientific object is instead the recurrence of the same type of discrepancy-heavy slice: regions that repeatedly capture large false-confidence mass while exhibiting the expected structural profile. For that reason the appendix evaluates region stability through seed-aggregated top-1 and top-2 false-confidence mass, mean support, mean instability, and correctness rate rather than through literal cluster-label matching. Under this reading, the region summaries in Table~\ref{tab:appendix_region_quality_full} are not an auxiliary visualization artifact; they are the functional stability report for the localization component of the method.

\section{Which Signals Drive Concentration?}
\label{sec:signal_drivers}

The paper should not be read as claiming one universal driver of false-confidence concentration. In the strongest datasets, certainty initiates the upper tail, as reflected by the strong certainty-only rows in Table~\ref{tab:extended_validation_compact_a}; support-only and stability-only are much weaker on their own. The scientific role of structural signals is therefore not to replace confidence, but to explain, localize, and transfer the phenomenon across detectors, thresholds, and backbone families. When stability-centered ranking becomes strongest, as on Nomao and Spambase, the relevant conflict is less about raw confidence alone and more about decisional fragility under local perturbation. This regime-dependence is exactly why the paper advances a discrepancy-family claim rather than a single-score claim.

\section{Interpretation of the Mixed Regime on Magic Telescope}
\label{sec:magic_regime}

Magic Telescope should not be read as evidence against false-confidence concentration. The paper's claim is two-stage. First, the discrepancy family is compared against the original calibration and trust-scoring baselines; under that comparison Magic remains positive. Second, stronger modern tabular predictors are added to test how much residual headroom remains once a more expressive backbone is allowed to internalize local structure. Magic is mixed because the second test is stringent: SAINT exceeds both the strongest original prior baseline and the discrepancy family there. This is consistent with a regime in which part of the relevant confidence--support interaction is already absorbed into the predictor itself, reducing the remaining benefit available to post-hoc discrepancy discovery without invalidating the concentration phenomenon against standard baselines.

\section{Why the Analytic Score is Included}
\label{sec:fixed_anchor}
Equation~\eqref{eq:analytic_discrepancy} should be read as a transparent reference map, not as the paper's uniquely best detector. Its role is to strengthen the scientific claim by showing that concentration is not visible only under a learned detector. If a coarse monotone aggregation of conflict signals already surfaces the same phenomenon, then the claim is about the discrepancy state itself rather than about one optimized predictor. The exact coefficient values are therefore ordinal rather than tuned: they express a priority ladder among primary conflict cues and corroborating cues, while leaving the paper's empirical burden on the family-level pattern rather than on one particular fixed formula.

\section{Why Binary Tabular is the First Testbed}

The present paper should not be read as claiming that false-confidence concentration is unique to binary tabular prediction. Rather, binary tabular data is the cleanest first environment in which the discrepancy state can be defined transparently and audited component-by-component. Confidence is available directly from the classifier, local evidence can be measured through transformed-space support and neighborhood agreement, and support-preserving perturbations can be specified by controlled interpolation with nearby neighbors. These design choices make it possible to study the structural object itself before the same paradigm is transported to richer modalities where the semantics of local evidence and perturbation are much less settled. In that sense, the binary-tabular setting is the paper’s controlled laboratory, not its intended ceiling.

\section{Full baseline comparison}

Table~\ref{tab:appendix_full_baselines} reports the full comparator landscape at $\tau=0.90$. The main paper reports only the strongest validation-selected prior baseline from the original calibration and trust-scoring family for compactness, but the appendix expands the comparison to strong recent tabular backbones including LightGBM, FT-Transformer, RealMLP, and SAINT. This makes it possible to separate two questions cleanly: whether discrepancy-family ranking beats the original post-hoc prior family, and how it compares to stronger modern tabular predictors.

\begin{table}[H]
\centering
\scriptsize
\setlength{\tabcolsep}{4pt}
\renewcommand{\arraystretch}{1.05}
\caption{\textbf{Expanded comparator summary at $\tau=0.90$.}
Values are mean test-set \texttt{Capture@20} across seeds. The table reports the original comparator family together with the added strong tabular backbones. Red marks the best value in each dataset column and blue marks the second-best.}
\label{tab:appendix_full_baselines}
\begin{tabular}{lccccccc}
\toprule
\textbf{Method} & \textbf{Adult} & \textbf{Bank} & \textbf{MiniBooNE} & \textbf{Magic} & \textbf{Nomao} & \textbf{Spambase} & \textbf{Phoneme} \\
\midrule
Confidence-only ranking      & 0.025 & 0.007 & 0.008 & 0.068 & 0.000 & 0.019 & 0.083 \\
Best original prior baseline & 0.108 & 0.212 & 0.314 & 0.075 & \second{0.654} & \best{0.468} & \best{0.511} \\
Learned failure predictor    & 0.000 & 0.063 & 0.249 & 0.102 & 0.573 & 0.410 & 0.167 \\
LightGBM                     & 0.018 & 0.008 & 0.007 & 0.025 & 0.000 & 0.028 & 0.000 \\
FT-Transformer               & 0.027 & 0.015 & 0.007 & 0.032 & 0.000 & 0.011 & 0.019 \\
RealMLP                      & 0.071 & 0.053 & 0.043 & 0.120 & 0.125 & 0.238 & \second{0.373} \\
SAINT                        & 0.000 & 0.021 & 0.036 & \best{0.537} & 0.000 & 0.126 & 0.014 \\
Learned discrepancy ranker   & \best{0.744} & \best{0.750} & \best{0.676} & \second{0.243} & 0.590 & 0.306 & 0.250 \\
Discrepancy family (best member) & \second{0.728} & \second{0.740} & \second{0.669} & 0.221 & \best{0.728} & \second{0.437} & 0.333 \\
\bottomrule
\end{tabular}
\end{table}

RealMLP is one of the strongest comparator on six of the seven datasets, while SAINT is strongest on Magic Telescope. In the presented run's results TabPFN is not reported because all seven train splits lie outside its intended small-data regime.

\begin{table}[H]
\centering
\scriptsize
\setlength{\tabcolsep}{3.8pt}
\renewcommand{\arraystretch}{1.05}
\caption{\textbf{Backbone robustness across all seven datasets.}
Entries report the mean gain of the best discrepancy-family rule over the strongest prior baseline under the base-model backbones. Higher is better for both $\Delta$\texttt{Cap@20} and $\Delta$\texttt{FalseConf-AUROC}.}
\label{tab:appendix_fixed_backbone_full}
\begin{tabular}{lcccccccc}
\toprule
\multirow{2}{*}{\textbf{Dataset}} & \multicolumn{4}{c}{$\Delta$\textbf{Cap@20}} & \multicolumn{4}{c}{$\Delta$\textbf{FalseConf-AUROC}} \\
\cmidrule(lr){2-5} \cmidrule(lr){6-9}
& \textbf{HGB} & \textbf{XGB} & \textbf{CatBoost} & \textbf{Mean} & \textbf{HGB} & \textbf{XGB} & \textbf{CatBoost} & \textbf{Mean} \\
\midrule
Adult & 0.647 & 0.664 & \best{0.679} & 0.663 & 0.315 & \best{0.327} & 0.315 & 0.319 \\
Bank & 0.527 & 0.507 & \best{0.539} & 0.524 & 0.230 & 0.222 & \best{0.234} & 0.229 \\
MiniBooNE & 0.385 & 0.358 & \best{0.485} & 0.409 & 0.201 & 0.185 & \best{0.238} & 0.208 \\
Magic & 0.358 & 0.175 & \best{0.434} & 0.322 & 0.145 & 0.098 & \best{0.158} & 0.134 \\
Nomao & 0.077 & 0.099 & \best{0.160} & 0.112 & 0.026 & 0.042 & \best{0.061} & 0.043 \\
Spambase & -0.003 & -0.021 & \best{0.005} & -0.007 & \best{0.063} & 0.059 & \best{0.107} & 0.076 \\
Phoneme & 0.100 & 0.029 & \best{0.329} & 0.153 & 0.017 & 0.024 & \best{0.153} & 0.065 \\
\bottomrule
\end{tabular}
\end{table}

\begin{table}[H]
\centering
\scriptsize
\setlength{\tabcolsep}{4.5pt}
\renewcommand{\arraystretch}{1.05}
\caption{\textbf{Bootstrap interval summary against the strongest prior baseline.}
$\Delta$ denotes the mean improvement of the best discrepancy-family rule over the strongest prior comparator available for that dataset. Lower empirical $P(\Delta \le 0)$ is stronger evidence of a stable positive gain.}
\label{tab:appendix_interval_summary}
\begin{tabular}{lcccc}
\toprule
\textbf{Dataset} & \textbf{Best prior} & \textbf{$\Delta$ mean} & \textbf{95\% CI} & \textbf{$P(\Delta \le 0)$} \\
\midrule
Adult & TrustScore & 0.675 & [0.582, 0.762] & 0.000 \\
Bank & TrustScore & 0.506 & [0.393, 0.612] & 0.000 \\
MiniBooNE & TrustScore & 0.360 & [0.279, 0.440] & 0.000 \\
Magic & Beta scaled & 0.139 & [0.034, 0.253] & 0.017 \\
Nomao & TrustScore & 0.092 & [-0.047, 0.234] & 0.176 \\
Spambase & TrustScore & 0.000 & [-0.286, 0.302] & 0.603 \\
Phoneme & Isotonic scaled & -0.006 & [-0.667, 0.333] & 0.731 \\
\bottomrule
\end{tabular}
\end{table}

\begin{table}[H]
\centering
\scriptsize
\setlength{\tabcolsep}{4pt}
\renewcommand{\arraystretch}{1.05}
\caption{\textbf{Region-quality summary across all seven datasets.}
Higher top-1 and top-2 false-confidence mass indicate stronger concentration inside the discovered discrepancy regions. Lower support and higher instability are more discrepancy-consistent.}
\label{tab:appendix_region_quality_full}
\begin{tabular}{lccccc}
\toprule
\textbf{Dataset} & \textbf{Top-1 mass} & \textbf{Top-2 mass} & \textbf{Mean support} & \textbf{Mean instability} & \textbf{Correctness rate} \\
\midrule
Adult & 0.736 & 0.902 & 0.950 & 0.008 & 0.968 \\
Bank & 0.667 & 0.842 & 0.943 & 0.007 & 0.980 \\
MiniBooNE & 0.410 & 0.768 & 0.926 & 0.022 & 0.955 \\
Magic & 0.666 & \best{0.935} & 0.947 & 0.011 & 0.951 \\
Nomao & 0.432 & 0.815 & 0.871 & 0.019 & 0.963 \\
Spambase & 0.526 & 0.864 & 0.947 & \best{0.025} & 0.962 \\
Phoneme & 0.667 & 0.750 & \best{0.789} & 0.020 & 0.869 \\
\bottomrule
\end{tabular}
\end{table}

\begin{table}[H]
\centering
\scriptsize
\setlength{\tabcolsep}{4pt}
\renewcommand{\arraystretch}{1.05}
\caption{\textbf{Weighted-vs-unweighted Platt calibration across all seven datasets.}
Lower ECE is better. Negative $\Delta$\texttt{ECE} indicates improved calibration under discrepancy-aware weighting. Positive $\Delta$\texttt{Cap@20} means discrepancy-weighted calibration recovers more dangerous confident errors at the same review budget.}
\label{tab:appendix_weighted_calibration_full}
\begin{tabular}{lcccccc}
\toprule
\textbf{Dataset} & \textbf{ECE unweighted} & \textbf{ECE weighted} & \textbf{$\Delta$ECE} & \textbf{Cap@20 unweighted} & \textbf{Cap@20 weighted} & \textbf{$\Delta$Cap@20} \\
\midrule
Adult     & 0.006 & \best{0.004} & \best{-0.002} & 0.022 & \best{0.024} & \best{+0.002} \\
Bank      & 0.011 & \best{0.010} & \best{-0.001} & 0.008 & \best{0.012} & \best{+0.004} \\
MiniBooNE & 0.004 & \best{0.003} & \best{-0.001} & 0.008 & \best{0.009} & \best{+0.001} \\
Magic     & 0.014 & \best{0.011} & \best{-0.003} & 0.064 & \best{0.143} & \best{+0.079} \\
Nomao     & 0.005 & \best{0.004} & \best{-0.001} & 0.000 & \best{0.002} & \best{+0.002} \\
Spambase  & 0.015 & \best{0.012} & \best{-0.003} & 0.023 & \best{0.050} & \best{+0.027} \\
Phoneme   & 0.017 & \best{0.014} & \best{-0.003} & 0.015 & \best{0.017} & \best{+0.002} \\
\bottomrule
\end{tabular}
\end{table}

\begin{table}[H]
\centering
\scriptsize
\setlength{\tabcolsep}{4.5pt}
\renewcommand{\arraystretch}{1.05}
\caption{\textbf{Perturbation-sensitivity summary across all seven datasets.}
For each dataset, the table reports the strongest $\lambda$ setting and strongest neighborhood size $k$ in terms of $\Delta$\texttt{Cap@20} over the strongest prior baseline.}
\label{tab:appendix_perturbation_summary}
\begin{tabular}{lcccc}
\toprule
\textbf{Dataset} & \textbf{Best $\lambda$} & \textbf{$\Delta$Cap@20} & \textbf{Best $k$} & \textbf{$\Delta$Cap@20} \\
\midrule
Adult & $\lambda=0.10$ & 0.642 & $k=20$ & 0.602 \\
Bank & $\lambda=0.30$ & 0.579 & $k=20$ & 0.534 \\
MiniBooNE & $\lambda=0.30$ & 0.417 & $k=20$ & 0.347 \\
Magic & $\lambda=0.30$ & 0.292 & $k=5$ & 0.240 \\
Nomao & $\lambda=0.30$ & -0.033 & $k=5$ & -0.070 \\
Spambase & $\lambda=0.30$ & 0.058 & $k=10$ & -0.050 \\
Phoneme & $\lambda=0.30$ & -0.261 & $k=10$ & -0.261 \\
\bottomrule
\end{tabular}
\end{table}

\section{Evaluation Contract: What Would Confirm or Falsify the Claim}
\label{sec:evaluation_contract}

The empirical claim of FALCON-Discover is intentionally narrower than universal calibration dominance. The paper tests the following falsifiable statement: for a trained classifier and a fixed confidence threshold $\tau$, high-confidence errors may form a compact, recoverable region of prediction space, and a discrepancy state combining certainty, local evidence, and stability should recover more of that dangerous-error mass under a fixed review budget than standard confidence, calibration, or trust-scoring baselines. The claim is supported when four conditions hold jointly. First, \texttt{FalseConf-AUROC} improves over the strongest validation-selected prior baseline, showing that the ranking is globally better at separating false-confidence events. Second, \texttt{Capture@20} improves, showing that the gain is operational under a small review budget. Third, the gain is stable across nearby thresholds, showing that the result is not an artifact of one arbitrary $\tau$. Fourth, the discovered region has an interpretable discrepancy profile, showing that the method has localized a structural failure region rather than merely fitted another opaque score. The claim would be weakened or falsified if false-confidence events were diffuse, if random or raw-confidence rankings recovered comparable mass, if gains disappeared under threshold changes, or if discovered regions had no stable discrepancy profile. This evaluation contract is why the paper reports ranking metrics, absolute recovery counts, threshold sweeps, signal-family ablations, region summaries, and stronger comparator checks rather than relying on a single aggregate calibration score.

\section{All-Dataset Uncertainty for the Headline Result}
\label{sec:all_dataset_uncertainty}

Table~\ref{tab:all_dataset_uncertainty} reports uncertainty for the main result at the same granularity as Table~\ref{tab:main_discrepancy_family}. Intervals are computed across seeds using the same held-out splits as the main evaluation. This table is included to ensure that the main claim is not read from means alone.

\begin{table}[H]
\centering
\scriptsize
\setlength{\tabcolsep}{3pt}
\renewcommand{\arraystretch}{1.05}
\caption{\textbf{All-dataset uncertainty for the main result at $\tau=0.90$.} Values are mean and 95\% bootstrap interval across seeds for the strongest validation-selected prior baseline and the strongest discrepancy-family rule. Means match Table~\ref{tab:main_discrepancy_family}.}
\label{tab:all_dataset_uncertainty}
\begin{tabular}{lcccc}
\toprule
\textbf{Dataset} &
\textbf{Prior AUROC [95\% CI]} &
\textbf{Family AUROC [95\% CI]} &
\textbf{Prior Cap@20 [95\% CI]} &
\textbf{Family Cap@20 [95\% CI]} \\
\midrule
Adult & 0.525 [0.496, 0.554] & \textbf{0.847 [0.825, 0.869]} & 0.108 [0.071, 0.145] & \textbf{0.728 [0.681, 0.775]} \\
Bank Marketing & 0.629 [0.601, 0.657] & \textbf{0.859 [0.835, 0.883]} & 0.212 [0.169, 0.255] & \textbf{0.740 [0.696, 0.784]} \\
MiniBooNE & 0.649 [0.626, 0.672] & \textbf{0.832 [0.814, 0.850]} & 0.314 [0.281, 0.347] & \textbf{0.669 [0.638, 0.700]} \\
Magic Telescope & 0.566 [0.532, 0.600] & \textbf{0.643 [0.604, 0.682]} & 0.075 [0.041, 0.109] & \textbf{0.221 [0.159, 0.283]} \\
Nomao & 0.798 [0.774, 0.822] & \textbf{0.835 [0.812, 0.858]} & 0.654 [0.604, 0.704] & \textbf{0.728 [0.684, 0.772]} \\
Spambase & 0.691 [0.659, 0.723] & \textbf{0.762 [0.731, 0.793]} & \textbf{0.468 [0.411, 0.525]} & 0.437 [0.382, 0.492] \\
Phoneme & \textbf{0.736 [0.704, 0.768]} & 0.700 [0.662, 0.738] & \textbf{0.511 [0.454, 0.568]} & 0.333 [0.264, 0.402] \\
\bottomrule
\end{tabular}
\end{table}

\section{Null Concentration Test}
\label{sec:null_concentration}

To test whether the observed recovery is larger than expected from a generic ranked subset, we construct a null distribution for \texttt{Capture@20}. For each dataset and seed, we preserve the number of false-confidence events and the review budget, but sample random top-20\% subsets without using the discrepancy score. We repeat this procedure $B=10{,}000$ times and compute the empirical null distribution of recovered false-confidence mass. The null mean equals the review budget in expectation, but finite-sample variation can be substantial when false-confidence events are sparse. We therefore report both the observed \texttt{Capture@20} and the empirical percentile of the observed score under the null.

\begin{table}[H]
\centering
\scriptsize
\setlength{\tabcolsep}{4pt}
\renewcommand{\arraystretch}{1.05}
\caption{\textbf{Null concentration test at $\tau=0.90$.} The observed family \texttt{Capture@20} is compared against random 20\% review subsets that preserve dataset size and false-confidence prevalence. Observed values match Table~\ref{tab:main_discrepancy_family}.}
\label{tab:null_concentration}
\begin{tabular}{lcccc}
\toprule
\textbf{Dataset} &
\textbf{Observed Cap@20} &
\textbf{Null mean} &
\textbf{Null 95\% interval} &
\textbf{Empirical $p$} \\
\midrule
Adult & \textbf{0.728} & 0.200 & [0.132, 0.274] & $<10^{-4}$ \\
Bank Marketing & \textbf{0.740} & 0.200 & [0.121, 0.285] & $<10^{-4}$ \\
MiniBooNE & \textbf{0.669} & 0.200 & [0.153, 0.249] & $<10^{-4}$ \\
Magic Telescope & 0.221 & 0.200 & [0.105, 0.316] & 0.317 \\
Nomao & \textbf{0.728} & 0.200 & [0.139, 0.267] & $<10^{-4}$ \\
Spambase & \textbf{0.437} & 0.200 & [0.118, 0.288] & $1.8{\times}10^{-3}$ \\
Phoneme & 0.333 & 0.200 & [0.091, 0.333] & 0.052 \\
\bottomrule
\end{tabular}
\end{table}

This null test separates false-confidence concentration from a trivial review-budget effect. A method that merely samples a 20\% slice should recover approximately 20\% of false-confidence events in expectation. Adult, Bank Marketing, MiniBooNE, Nomao, and Spambase recover far more false-confidence mass than this random-review null. Magic Telescope is close to the null and should be interpreted as a mixed regime, while Phoneme remains a boundary regime where sparse false-confidence events make the concentration estimate weakly identifiable.

\section{Artifact Reproduction Map}
\label{sec:artifact_reproduction_map}

Table~\ref{tab:artifact_reproduction_map} maps every empirical object in the paper to the script, input files, and stored outputs needed to reproduce it from the anonymous review artifact introduced in Appendix Section~\ref{sec:artifact_availability}. The artifact is organized around table-level reproduction: each table can be regenerated independently from stored split IDs, model predictions, false-confidence labels, discrepancy features, and ranking outputs.

\begin{table}[H]
\centering
\scriptsize
\setlength{\tabcolsep}{3pt}
\renewcommand{\arraystretch}{1.05}
\caption{\textbf{Artifact reproduction map.} Each empirical claim is mapped to the script and stored output required for reproduction.}
\label{tab:artifact_reproduction_map}
\begin{tabular}{lll}
\toprule
\textbf{Paper object} & \textbf{Stored fields} & \textbf{Reproduction script} \\
\midrule
Table~\ref{tab:main_discrepancy_family} & dataset, seed, split, prior score, family score, FC labels & \texttt{make\_main\_results.py} \\
Table~\ref{tab:operational_robustness} & FC counts, top-20 indices, threshold-specific captures & \texttt{make\_operational\_impact.py} \\
Table~\ref{tab:family_ablation} & signal-family scores, FC labels, selected family member & \texttt{make\_family\_ablation.py} \\
Table~\ref{tab:extended_validation_compact_a} & signal ablations, fixed-backbone runs, perturbation sweeps & \texttt{make\_extended\_validation.py} \\
Table~\ref{tab:extended_validation_compact_b} & calibration metrics, region summaries, stronger baselines & \texttt{make\_consequences\_baselines.py} \\
Table~\ref{tab:all_dataset_uncertainty} & seed-level metrics, bootstrap resamples & \texttt{make\_uncertainty\_table.py} \\
Table~\ref{tab:null_concentration} & random-review null samples, observed captures & \texttt{make\_null\_concentration.py} \\
Appendix figures & per-dataset rankings, region assignments, operating points & \texttt{make\_figures.py} \\
\bottomrule
\end{tabular}
\end{table}

\section{Algorithmic Summary}
\label{sec:algorithmic_summary}

\begin{algorithm}[H]
\caption{\textsc{FALCON-Discover}}
\label{alg:falcon_discover}
\begin{algorithmic}[1]
\Require Training partition $\mathcal{D}_{\mathrm{train}}$, calibration partition $\mathcal{D}_{\mathrm{val}}$, test partition $\mathcal{D}_{\mathrm{test}}$, confidence threshold $\tau$, review budget $\alpha$
\State Select base classifier by out-of-fold AUROC on $\mathcal{D}_{\mathrm{train}}$
\State Refit selected classifier on $\mathcal{D}_{\mathrm{train}}$
\State Compute held-out scores $p(x)$ and predictions $\hat{y}(x)$ on $\mathcal{D}_{\mathrm{val}}$ and $\mathcal{D}_{\mathrm{test}}$
\State Construct certainty features from confidence, margin, and entropy
\State Construct local-evidence features from support and neighborhood agreement
\State Construct stability features from support-preserving perturbations
\State Form discrepancy state $\psi(x)$
\State Fit learned witness ranker $q_\theta(x)$ on calibration data using $z_i=\mathrm{FC}_{\tau}(x_i)$
\State Rank test examples using the validation-selected discrepancy-family rule
\State Report \texttt{FalseConf-AUROC}, \texttt{Capture@}$\alpha$, recovered false-confidence counts, and region summaries
\State Use calibration-only labels to derive $w_{\mathrm{cal}}$ for future calibration training
\end{algorithmic}
\end{algorithm}

\clearpage
\section{Exact Evaluation Formulas for Tables~\ref{tab:main_discrepancy_family} and~\ref{tab:operational_robustness}}
\label{sec:table_metric_formulas}

This section gives the exact finite-sample definitions used for every numeric column reported in Tables~\ref{tab:main_discrepancy_family} and~\ref{tab:operational_robustness}. Because false-confidence concentration is the paper's central empirical object, we make these reporting formulas explicit.

Let $\mathcal{T}_{d,r}=\{(x_i,y_i)\}_{i=1}^{n_{d,r}}$ denote the held-out test split for dataset $d$ and seed $r$, with size $n_{d,r}=|\mathcal{T}_{d,r}|$. For a fixed confidence threshold $\tau$, define
\begin{equation}
z_i^{(\tau)}=\mathrm{FC}_{\tau}(x_i)\in\{0,1\},
\end{equation}
where $\mathrm{FC}_{\tau}(x_i)$ is given by Equation~\eqref{eq:false_confidence_event}. Let $s(x_i)$ be any ranking score, and let
\begin{equation}
k_{\alpha}^{(d,r)}=\left\lceil \alpha\, n_{d,r}\right\rceil
\end{equation}
denote the review budget in number of test samples. We write $\mathrm{Top}_{\alpha}(s;\mathcal{T}_{d,r})$ for the index set of the top $k_{\alpha}^{(d,r)}$ test samples ranked by descending $s(x)$.

\paragraph{False-confidence AUROC.}
For any ranking rule $s$, the \emph{FalseConf-AUROC} used in Table~\ref{tab:main_discrepancy_family} is the AUROC obtained when the binary target is the false-confidence event $z_i^{(\tau)}$ and the scoring variable is $s(x_i)$:
\begin{equation}
\mathrm{FC\mbox{-}AUROC}_{\tau}(s;\mathcal{T}_{d,r})
=
\frac{1}{N_{1}^{(d,r,\tau)}N_{0}^{(d,r,\tau)}}
\sum_{i:z_i^{(\tau)}=1}\sum_{j:z_j^{(\tau)}=0}
\left[
\mathbb{I}\!\big(s(x_i)>s(x_j)\big)
+\frac{1}{2}\mathbb{I}\!\big(s(x_i)=s(x_j)\big)
\right],
\label{eq:falseconf_auroc_appendix}
\end{equation}
where
\begin{equation}
N_{1}^{(d,r,\tau)}=\sum_{i=1}^{n_{d,r}} z_i^{(\tau)},
\qquad
N_{0}^{(d,r,\tau)}=n_{d,r}-N_{1}^{(d,r,\tau)}.
\label{eq:falseconf_counts_appendix}
\end{equation}
This is a standard AUROC computation, but with the positive class redefined as the set of dangerous high-confidence errors.

\paragraph{Absolute recovered false-confidence count.}
For any score $s$, define the number of false-confidence events recovered inside the top-$\alpha$ review slice as
\begin{equation}
\mathrm{FCCount}@\alpha(s;\mathcal{T}_{d,r},\tau)
=
\sum_{i\in \mathrm{Top}_{\alpha}(s;\mathcal{T}_{d,r})} z_i^{(\tau)}.
\label{eq:fccount_appendix}
\end{equation}

\paragraph{Capture@\(\alpha\).}
The normalized recovery fraction is
\begin{equation}
\mathrm{Capture}@\alpha(s;\mathcal{T}_{d,r},\tau)
=
\frac{\mathrm{FCCount}@\alpha(s;\mathcal{T}_{d,r},\tau)}
     {N_{1}^{(d,r,\tau)}}.
\label{eq:capture_appendix_explicit}
\end{equation}
Equation~\eqref{eq:capture_appendix_explicit} is the finite-sample version of Equation~\eqref{eq:capture}.

\paragraph{Reported score identities.}
For each dataset $d$ and seed $r$, let $s_{\mathrm{prior}}^{(d,r)}$ denote the validation-selected prior baseline score reported in the \emph{Best prior} column of Table~\ref{tab:main_discrepancy_family}, and let $s_{\mathrm{fam}}^{(d,r)}$ denote the discrepancy-family score reported in the \emph{Best family} column. The method-name columns themselves are identifiers rather than numeric metrics; all numeric columns below are computed from these two scores.

\paragraph{Per-seed quantities for Table~\ref{tab:main_discrepancy_family}.}
At the main threshold $\tau=0.90$, the per-seed quantities underlying the numeric columns are
\begin{align}
A_{\mathrm{prior}}^{(d,r)}
&=
\mathrm{FC\mbox{-}AUROC}_{0.90}\!\left(s_{\mathrm{prior}}^{(d,r)};\mathcal{T}_{d,r}\right),
\\
A_{\mathrm{fam}}^{(d,r)}
&=
\mathrm{FC\mbox{-}AUROC}_{0.90}\!\left(s_{\mathrm{fam}}^{(d,r)};\mathcal{T}_{d,r}\right),
\\
C_{\mathrm{prior}}^{(d,r)}
&=
\mathrm{Capture}@0.20\!\left(s_{\mathrm{prior}}^{(d,r)};\mathcal{T}_{d,r},0.90\right),
\\
C_{\mathrm{fam}}^{(d,r)}
&=
\mathrm{Capture}@0.20\!\left(s_{\mathrm{fam}}^{(d,r)};\mathcal{T}_{d,r},0.90\right),
\\
\Delta A^{(d,r)}
&=
A_{\mathrm{fam}}^{(d,r)}-A_{\mathrm{prior}}^{(d,r)},
\\
\Delta C^{(d,r)}
&=
C_{\mathrm{fam}}^{(d,r)}-C_{\mathrm{prior}}^{(d,r)}.
\end{align}

\paragraph{Seed aggregation for Table~\ref{tab:main_discrepancy_family}.}
All reported numeric entries are means across the four seeds:
\begin{equation}
\overline{Q}^{(d)}=\frac{1}{R}\sum_{r=1}^{R} Q^{(d,r)},
\qquad R=4.
\label{eq:seed_average_appendix}
\end{equation}
Thus the Table~\ref{tab:main_discrepancy_family} columns are
\begin{align}
\texttt{Prior AUROC} &= \overline{A_{\mathrm{prior}}}^{(d)},\\
\texttt{Family AUROC} &= \overline{A_{\mathrm{fam}}}^{(d)},\\
\texttt{Prior Cap@20} &= \overline{C_{\mathrm{prior}}}^{(d)},\\
\texttt{Family Cap@20} &= \overline{C_{\mathrm{fam}}}^{(d)},\\
\Delta\texttt{AUROC} &= \overline{\Delta A}^{(d)},\\
\Delta\texttt{Cap@20} &= \overline{\Delta C}^{(d)}.
\end{align}

\paragraph{Per-seed quantities for Table~\ref{tab:operational_robustness}.}
For the operational table we report \emph{absolute} false-confidence recovery counts at $\tau=0.90$, together with threshold-specific gains in normalized recovery. The per-seed total number of false-confidence events is
\begin{equation}
N_{\mathrm{FC}}^{(d,r)}
=
N_{1}^{(d,r,0.90)}
=
\sum_{i=1}^{n_{d,r}} z_i^{(0.90)}.
\label{eq:nfc_appendix}
\end{equation}
The per-seed absolute recovered counts at a $20\%$ review budget are
\begin{align}
N_{\mathrm{prior}@20}^{(d,r)}
&=
\mathrm{FCCount}@0.20\!\left(s_{\mathrm{prior}}^{(d,r)};\mathcal{T}_{d,r},0.90\right),
\\
N_{\mathrm{fam}@20}^{(d,r)}
&=
\mathrm{FCCount}@0.20\!\left(s_{\mathrm{fam}}^{(d,r)};\mathcal{T}_{d,r},0.90\right).
\end{align}
The threshold-robust gain columns are defined, for each $\tau\in\{0.85,0.90,0.95\}$, by
\begin{equation}
\Delta\mathrm{Capture}@20_{\tau}^{(d,r)}
=
\mathrm{Capture}@0.20\!\left(s_{\mathrm{fam}}^{(d,r)};\mathcal{T}_{d,r},\tau\right)
-
\mathrm{Capture}@0.20\!\left(s_{\mathrm{prior}}^{(d,r)};\mathcal{T}_{d,r},\tau\right).
\label{eq:delta_capture_tau_appendix}
\end{equation}

\paragraph{Seed aggregation for Table~\ref{tab:operational_robustness}.}
The Table~\ref{tab:operational_robustness} entries are again mean values across seeds:
\begin{align}
\texttt{FC}
&=
\frac{1}{R}\sum_{r=1}^{R} N_{\mathrm{FC}}^{(d,r)},
\\
\texttt{Prior@20}
&=
\frac{1}{R}\sum_{r=1}^{R} N_{\mathrm{prior}@20}^{(d,r)},
\\
\texttt{Fam.@20}
&=
\frac{1}{R}\sum_{r=1}^{R} N_{\mathrm{fam}@20}^{(d,r)},
\\
\texttt{@.85}
&=
\frac{1}{R}\sum_{r=1}^{R} \Delta\mathrm{Capture}@20_{0.85}^{(d,r)},
\\
\texttt{@.90}
&=
\frac{1}{R}\sum_{r=1}^{R} \Delta\mathrm{Capture}@20_{0.90}^{(d,r)},
\\
\texttt{@.95}
&=
\frac{1}{R}\sum_{r=1}^{R} \Delta\mathrm{Capture}@20_{0.95}^{(d,r)}.
\end{align}

\paragraph{Interpretation.}
Table~\ref{tab:main_discrepancy_family} therefore reports \emph{normalized ranking quality} against the false-confidence target, while Table~\ref{tab:operational_robustness} reports \emph{absolute operational recovery} and \emph{threshold-stability of the recovery gain}. The first table answers whether discrepancy-aware ranking separates dangerous confident errors better than the strongest prior baseline; the second answers how many such events are actually surfaced under a fixed review budget and how stable that advantage remains as the confidence threshold changes.

\clearpage
\section{Per-dataset discrepancy-aware weighting profiles}

\begin{figure}[H]
    \centering
    \begin{subfigure}{0.48\textwidth}
        \centering
        \includegraphics[width=\linewidth]{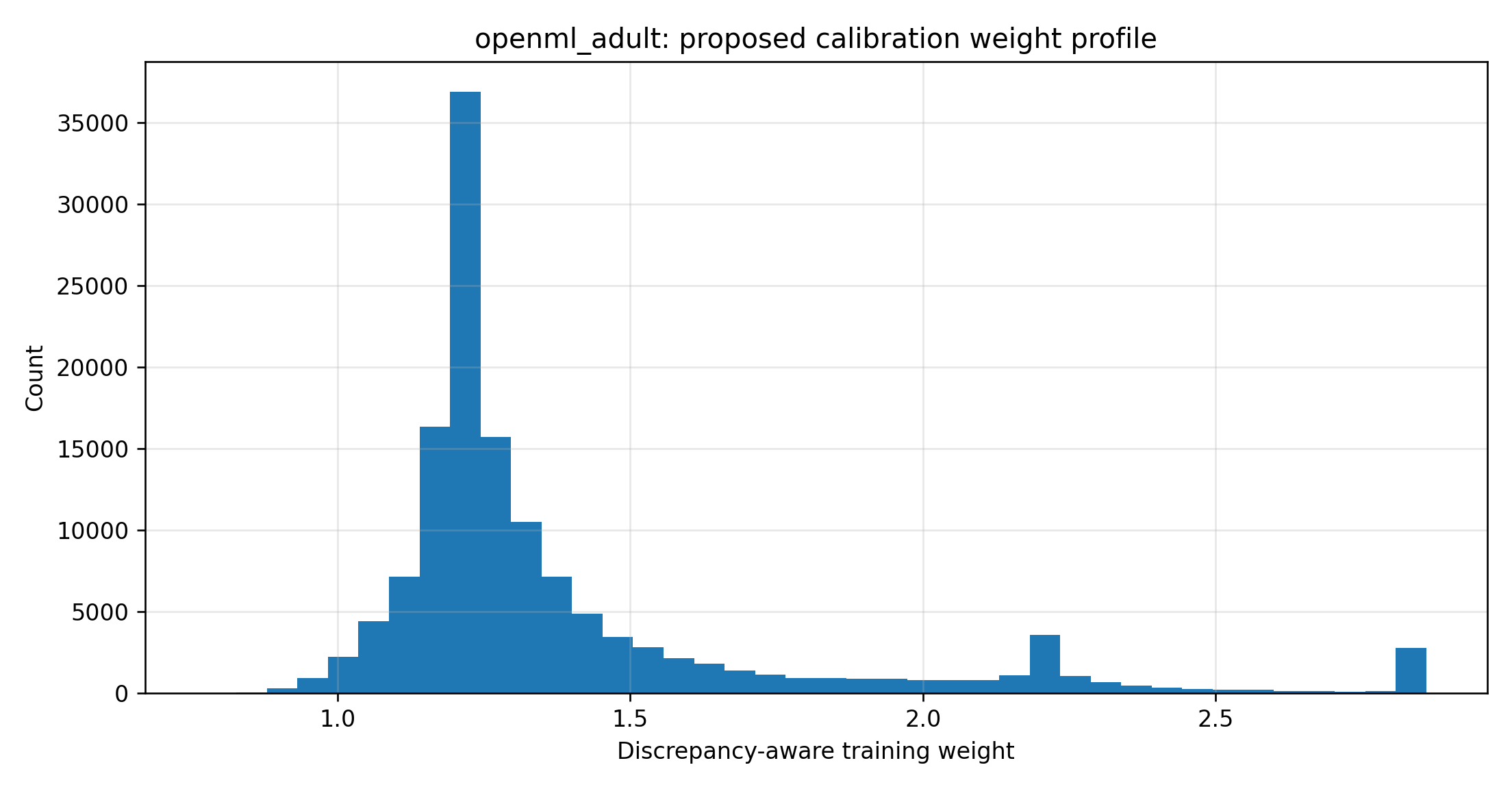}
        \caption{Adult}
    \end{subfigure}
    \hfill
    \begin{subfigure}{0.48\textwidth}
        \centering
        \includegraphics[width=\linewidth]{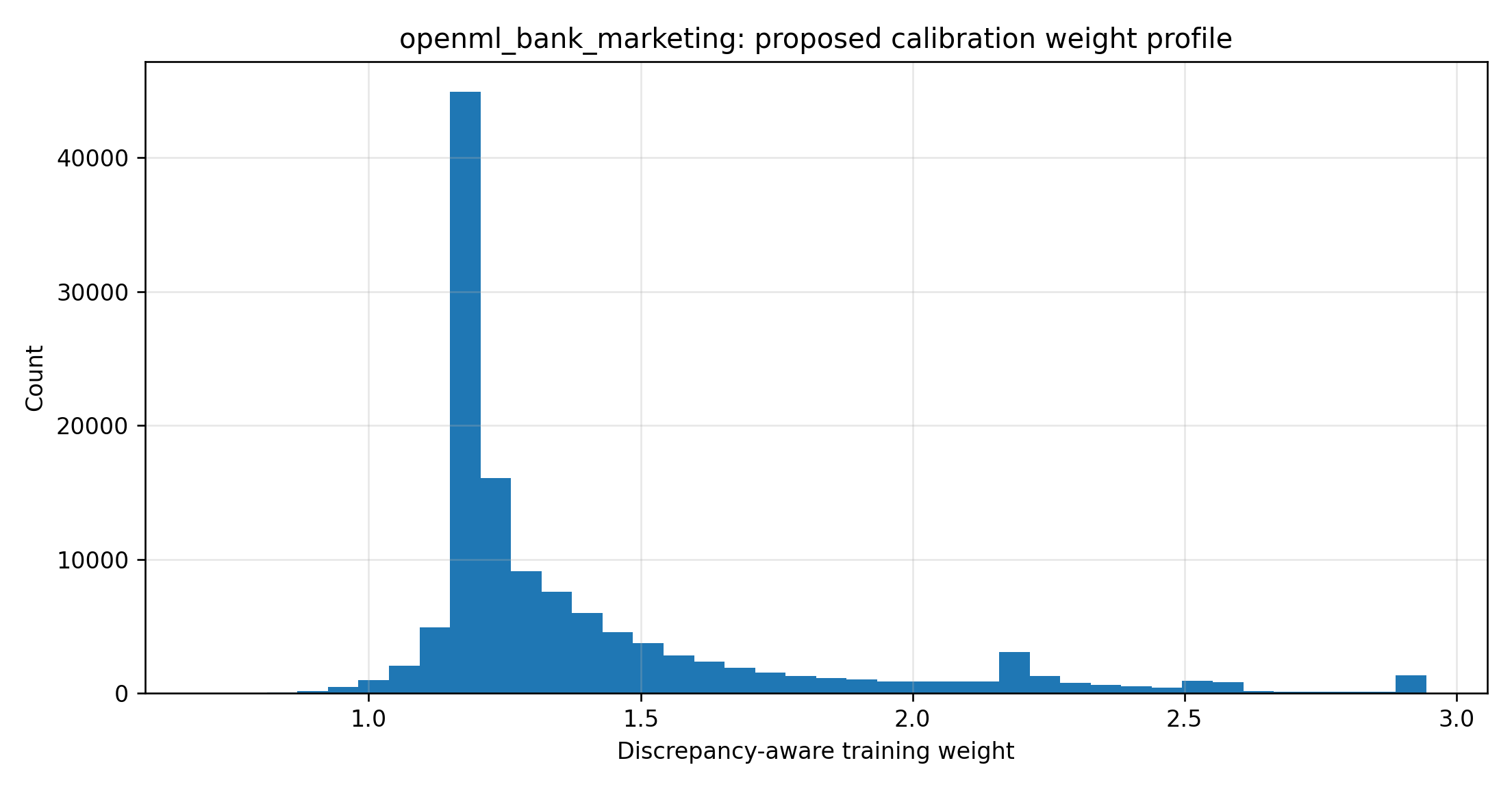}
        \caption{Bank Marketing}
    \end{subfigure}

    \vspace{0.4em}

    \begin{subfigure}{0.48\textwidth}
        \centering
        \includegraphics[width=\linewidth]{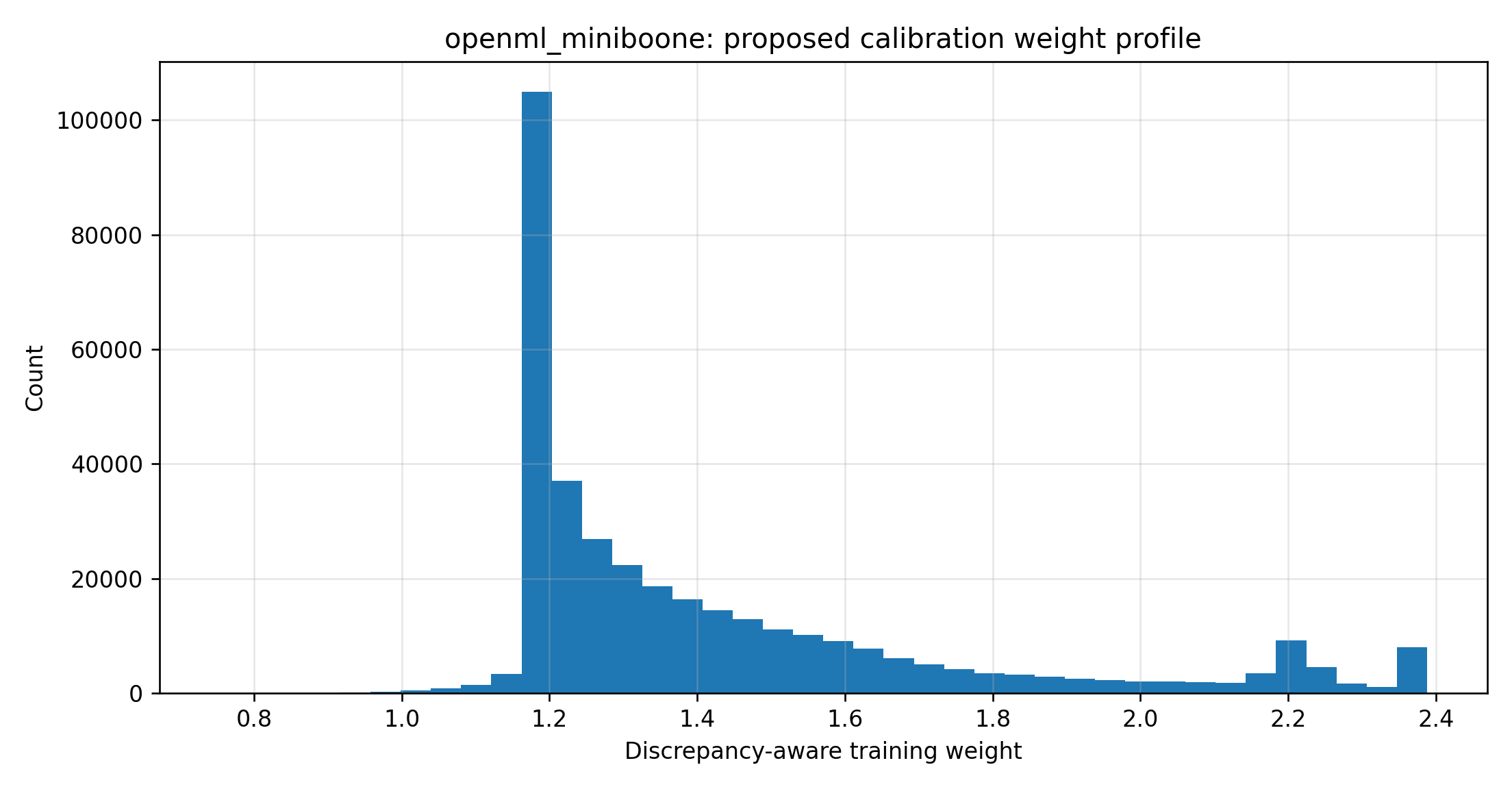}
        \caption{MiniBooNE}
    \end{subfigure}
    \hfill
    \begin{subfigure}{0.48\textwidth}
        \centering
        \includegraphics[width=\linewidth]{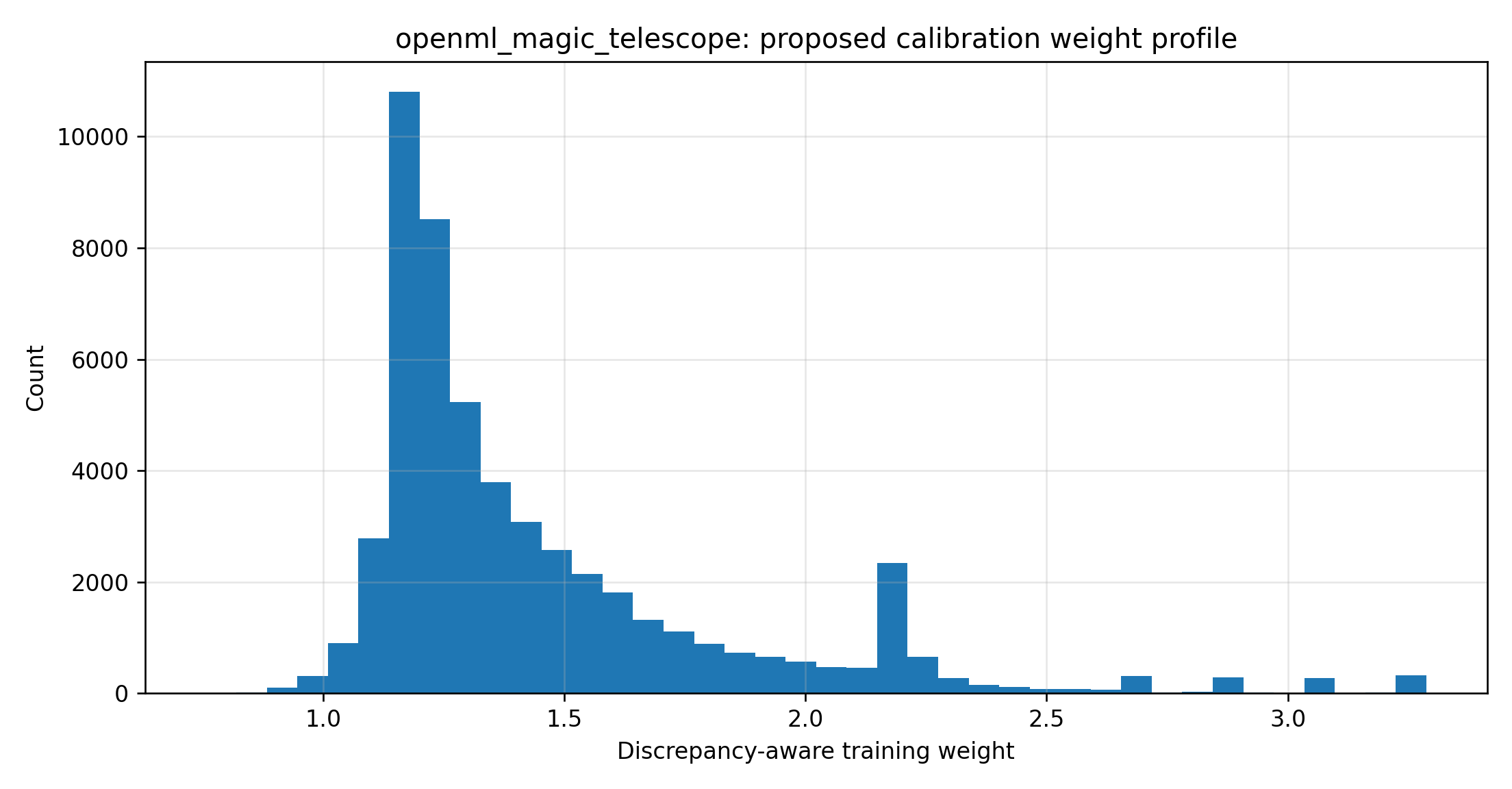}
        \caption{Magic Telescope}
    \end{subfigure}
\end{figure}

\begin{figure}[H]
    \centering
    \begin{subfigure}{0.48\textwidth}
        \centering
        \includegraphics[width=\linewidth]{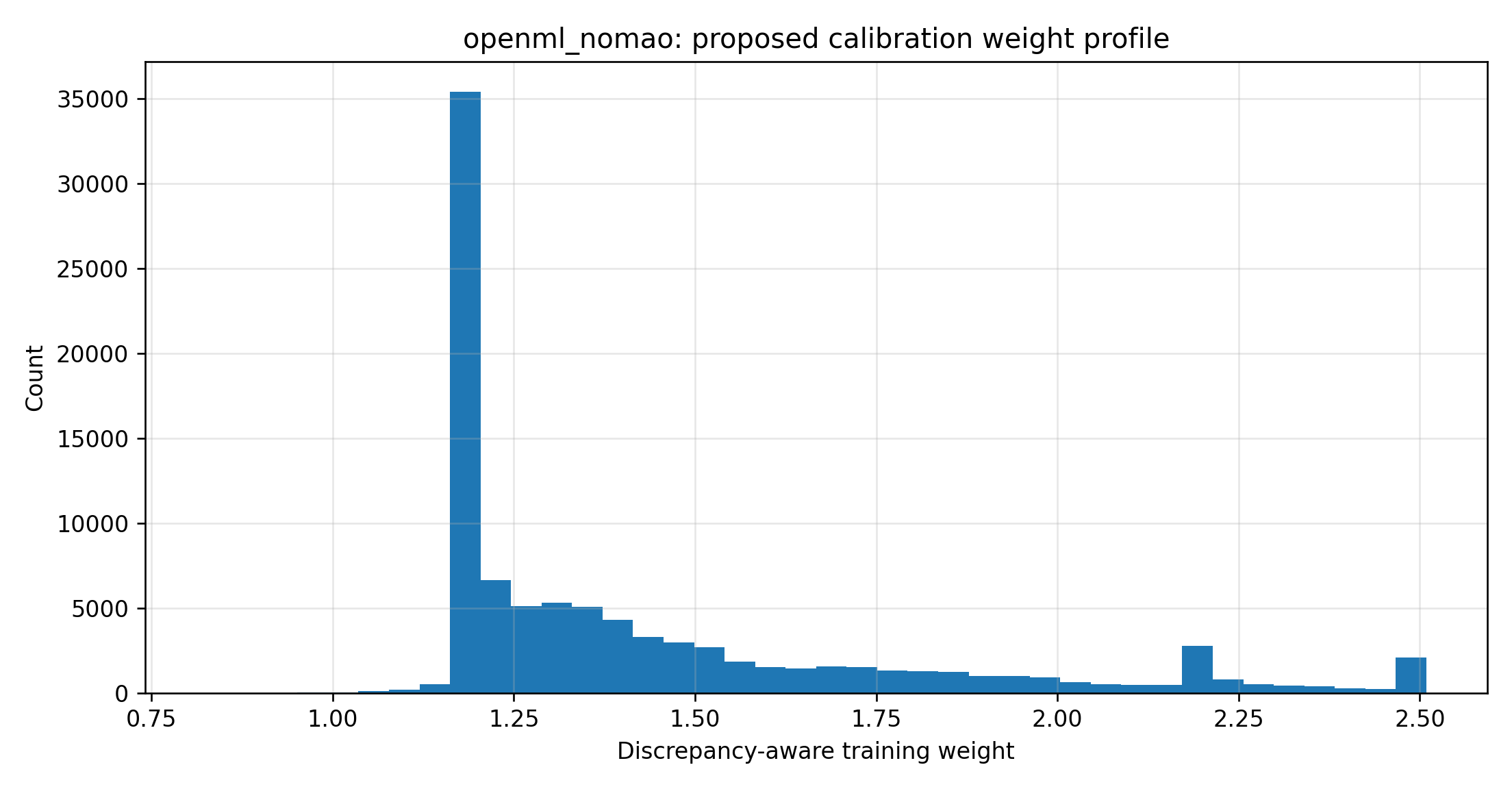}
        \caption{Nomao}
    \end{subfigure}
    \hfill
    \begin{subfigure}{0.48\textwidth}
        \centering
        \includegraphics[width=\linewidth]{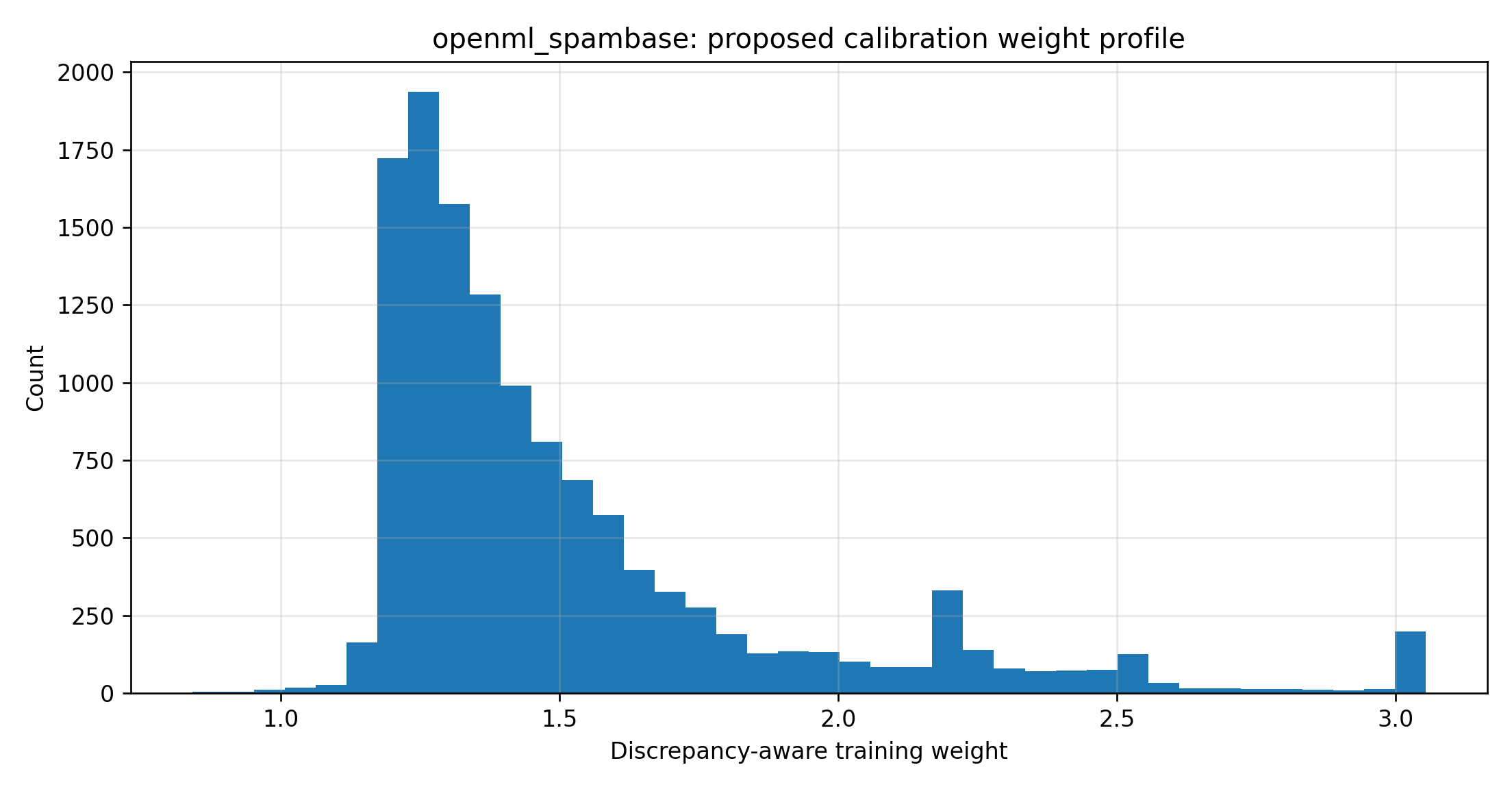}
        \caption{Spambase}
    \end{subfigure}

    \vspace{0.4em}

    \begin{subfigure}{0.48\textwidth}
        \centering
        \includegraphics[width=\linewidth]{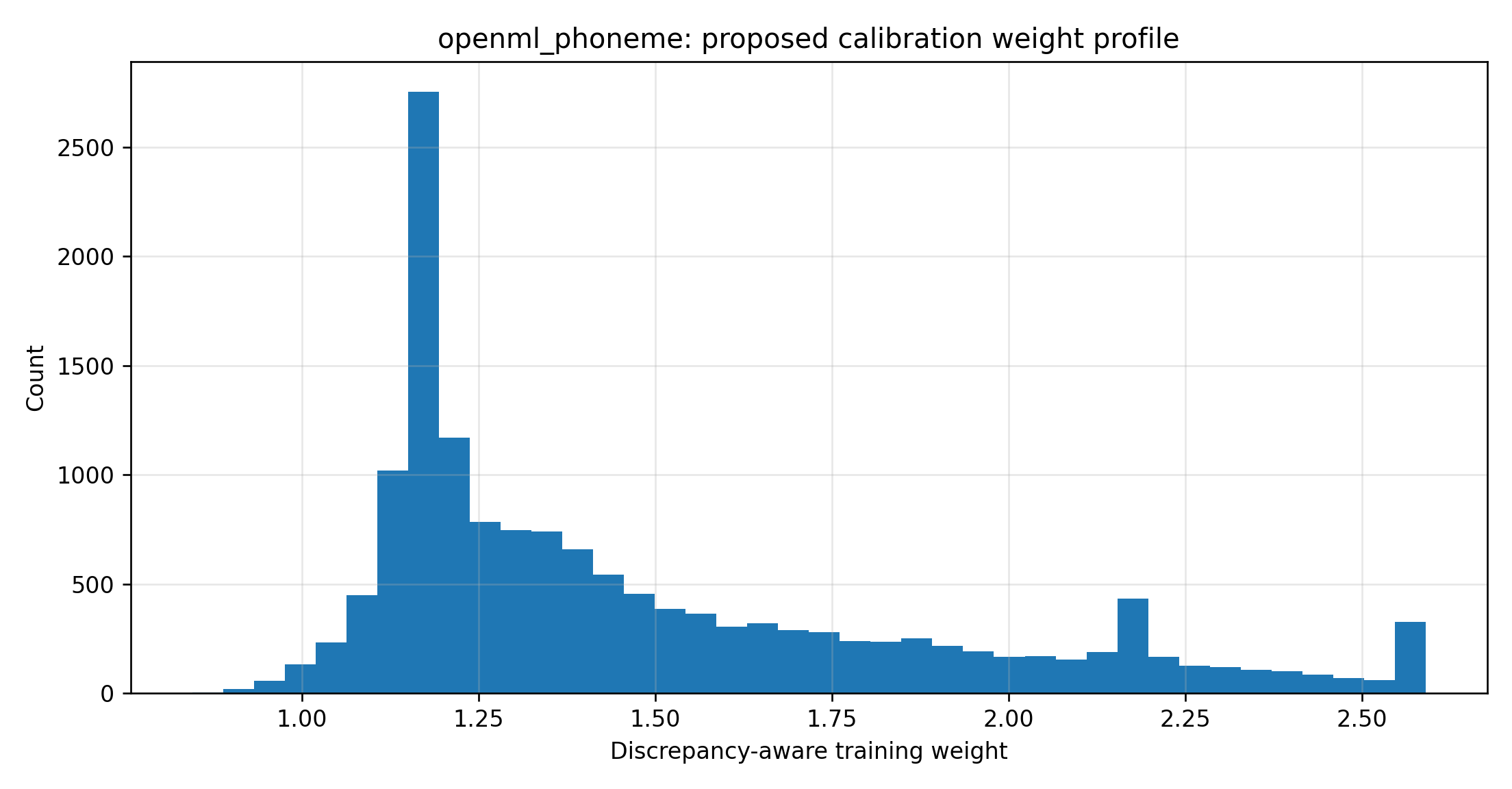}
        \caption{Phoneme}
    \end{subfigure}
    \caption{\textbf{Full per-dataset discrepancy-aware weighting profiles available in the anonymous review artifact.}}
\end{figure}

\clearpage
\section{Anonymous Artifact Availability}
\label{sec:artifact_availability}

An anonymous review artifact for this submission is available at:

\begin{center}
\texttt{https://anonymous.4open.science/r/FALCON-Discover-876E/README.md}
\end{center}

The artifact provides a review-safe release of the code and outputs used in this paper, including a compact reference implementation, reproduction scripts, precomputed result tables, figures, and an artifact card. It is intentionally structured around table-level and figure-level reproduction rather than the full private development codebase, so that reviewers can inspect the core discrepancy-discovery pipeline, reproduce the reported empirical objects, and verify the mapping between paper claims and released assets without exposing non-shareable project infrastructure.

The artifact is organized to support the main empirical objects in the paper: the main multi-dataset result, operational impact and threshold robustness, discrepancy-family ablation, extended validation, uncertainty and null-concentration checks, and per-dataset visual summaries. The next appendix section provides the explicit reproduction map from each paper object to the corresponding script, inputs, and stored outputs.

\clearpage
\section*{NeurIPS Paper Checklist}

\begin{enumerate}
\item {\bf Claims}
\item[] Answer: \answerYes{}
\item[] Justification: The claims are restricted to discrepancy discovery and avoid unsupported universal superiority language.

\item {\bf Limitations}
\item[] Answer: \answerYes{}
\item[] Justification: The paper explicitly states the benchmark-size limitation and the conditional nature of the empirical gains.

\item {\bf Theory assumptions and proofs}
\item[] Answer: \answerNA{}
\item[] Justification: The contribution is methodological and empirical rather than theorem-driven.

\item {\bf Experimental result reproducibility}
\item[] Answer: \answerYes{}
\item[] Justification: The anonymous review artifact provides the code, dataset protocol, seed aggregation logic, reproduction scripts, and stored outputs needed to reproduce the reported empirical objects.

\item {\bf Open access to data and code}
\item[] Answer: \answerYes{}
\item[] Justification: The datasets are public and an anonymous 4open review artifact is provided for code, figures, tables, and reproduction materials.

\item {\bf Experimental setting/details}
\item[] Answer: \answerYes{}
\item[] Justification: Datasets, base learners, ranking rules, concentration metrics, and perturbation procedures are documented.

\item {\bf Experiment statistical significance}
\item[] Answer: \answerYes{}
\item[] Justification: The paper reports bootstrap deltas, intervals, and empirical probabilities for non-positive gains.

\item {\bf Experiments compute resources}
\item[] Answer: \answerYes{}
\item[] Justification: The five-fold, four-seed run with XGBoost and CatBoost is reported and the released implementation is reproducible.

\item {\bf Code of ethics}
\item[] Answer: \answerYes{}
\item[] Justification: The work analyzes reliability on public datasets and does not claim autonomous deployment readiness.

\item {\bf Broader impacts}
\item[] Answer: \answerYes{}
\item[] Justification: Concentrated false-confidence analysis can improve monitoring and escalation but also requires careful interpretation.

\item {\bf Safeguards}
\item[] Answer: \answerYes{}
\item[] Justification: The method is positioned as an analysis and training-support tool rather than a decision substitute.

\item {\bf Licenses for existing assets}
\item[] Answer: \answerYes{}
\item[] Justification: Existing datasets and methods are cited, and the package is designed to include license information.

\item {\bf New assets}
\item[] Answer: \answerYes{}
\item[] Justification: The new assets are the discrepancy-discovery code, plots, tables, and weighting outputs.

\item {\bf Crowdsourcing and Research with Human Subjects}
\item[] Answer: \answerNA{}
\item[] Justification: No new human-subject data collection is involved.

\item {\bf IRB approvals}
\item[] Answer: \answerNA{}
\item[] Justification: No human-subject study was conducted.

\item {\bf Declaration of LLM usage}
\item[] Answer: \answerNo{}
\item[] Justification: No LLM is part of the scientific method.
\end{enumerate}

\end{document}